\documentclass{article}

\usepackage{iclr2027_conference,times}

\renewcommand\cite{\citep}

\usepackage{latexsym}
\usepackage[T1]{fontenc}
\usepackage[utf8]{inputenc}
\usepackage{url}
\usepackage{multirow}
\usepackage{floatrow}
\usepackage{amsmath}
\usepackage{amssymb}
\usepackage{booktabs}
\usepackage{enumitem}
\usepackage{pifont}
\usepackage{hyperref}
\usepackage{tcolorbox}
\tcbuselibrary{skins,breakable}
\usepackage{microtype}
\usepackage{inconsolata}
\usepackage{graphicx}
\usepackage{tikz}
\usepackage{xcolor}
\usepackage{wrapfig}
\usepackage{xurl}
\usepackage{needspace}

\newcommand{\scigenicon}{%
  \raisebox{-0.28em}{\includegraphics[height=1.05em]{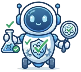}}%
  \hspace{0.1em}%
}

\title{\texorpdfstring{\scigenicon SciGen-Verifier: A Multimodal Reasoner\\ for Explainable Verification\\ in Scientific Image Generation}{SciGen-Verifier: A Multimodal Reasoner for Explainable Verification in Scientific Image Generation}}

\author{\textbf{Jiali Chen}$^{1,2}$, \textbf{Zhengteng Lin}$^{1}$, \textbf{Zuqi Wang}$^{1}$, \textbf{Shirong Lin}$^{1}$, \textbf{Xi Yu}$^{1}$,\\
\textbf{Xusen Hei}$^{1}$, \textbf{DingBa Fu}$^{1}$, \textbf{Jiayuan Xie}$^{2}$, \textbf{Yi Cai}$^{1}$\thanks{Corresponding author.}\\
\\
$^1$South China University of Technology \qquad $^2$The Hong Kong Polytechnic University\\
\texttt{gaalik.chen@connect.polyu.hk}}

\iclrfinalcopy

\newcommand{\ie}{\emph{i.e.}}
\newcommand{\eg}{\emph{e.g.}}

\begin{document}
\setcounter{footnote}{1}
\maketitle
\lhead{}

\begin{abstract}
In realistic education, a solution is often expressed not only in words but in a drawing---a circuit, a geometric construction, a function plot---and a teacher must grade the drawing as carefully as the text.
Recent advances in unified multimodal models have enabled scientific image generation, yet verifying the correctness of these specialized visual outputs remains a critical bottleneck: errors often arise from intricate domain knowledge, structural reasoning, and multi-step instruction rather than surface-level artifacts.
Existing verifiers mainly target natural images and compress judgement into scalar scores, leaving scientific coverage and explainable feedback for error correction underexplored.
To bridge this gap, we make three main contributions.
(1) We construct \textit{SciGen-Verify}, a benchmark dedicated to explainable verification of scientific image generation, spanning instruction following, multidisciplinary reasoning, and world knowledge domains. It contains a three-tier hierarchical protocol over the binary judgement, supporting explanation, and corrective editing instruction.
(2) We develop SciGen-Verifier, a reasoning-driven multimodal verifier trained via cold-start supervised fine-tuning followed by a curriculum-based two-stage reinforcement learning pipeline. The rubric-guided process rewards first strengthen scientific reasoning exploration and outcome rewards subsequently align output with ground-truth annotation.
(3) On \textit{SciGen-Verify}, SciGen-Verifier achieves competitive performance against much larger proprietary models. It further serves as a practical online critic for iterative image rectification.
\end{abstract}

\section{Introduction}
The field of multimodal models is undergoing a remarkable revolution, with vision-language models \cite{glm-5v-turbo,kimi-k2.5,seed2.0} tackling complex visual reasoning and generative models (\ie, unified multimodal models \cite{bagel,mmada,omnigen2} and diffusion-based generators \cite{dalle3,sd3}) enabling high-fidelity visual synthesis.
Recent advances in generative models are pushing text-to-image synthesis beyond open-domain natural scenes toward more specialized scientific image generation \cite{genexam, proimagebench}.
In educational scenarios, some scientific questions are answered by drawing rather than writing: sketching a circuit, constructing a geometric figure, or plotting a function.
As generative models begin to ``do homework'' in this visual form, a natural question follows: who grades these drawings, and how?
Unlike natural images, scientific images demand strict adherence to domain knowledge, structural reasoning, and multi-step instruction, where subtle deviations can cause the output to be scientifically invalid.
Reliable verification of these specialized visual outputs thus emerges as a critical bottleneck for a trustworthy scientific image generation system.

\begin{figure*}[t]
  \centering
  \includegraphics[width=\textwidth]{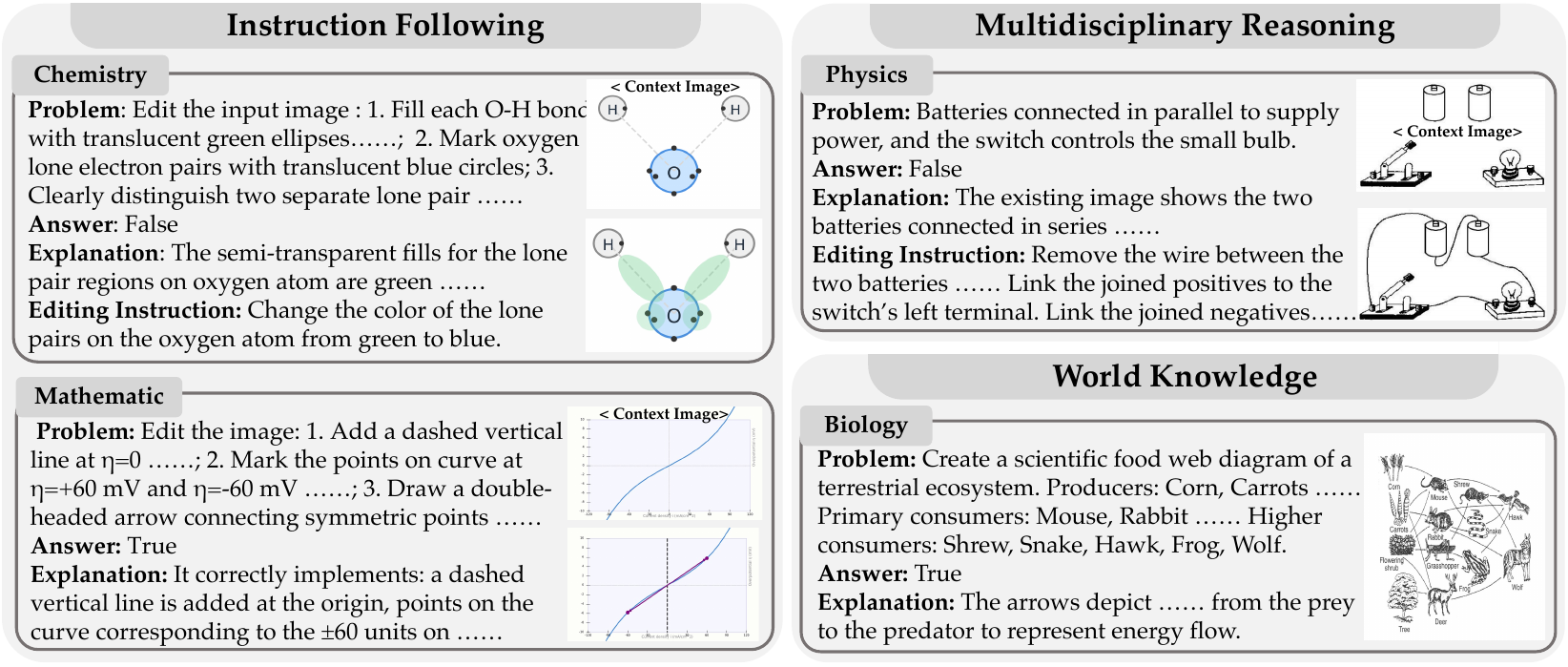}
  \caption{Overview of \textit{SciGen-Verify} benchmark across instruction following, multidisciplinary reasoning and world knowledge domains. The context image is the input image of the problem.
  }
  \label{fig:intro_case}
\end{figure*}

As illustrated in Fig.~\ref{fig:intro_case}, errors in scientific image generation fundamentally differ from surficial artifacts in natural images: they manifest as violated physical or geometric constraints, missing critical components, and subtle misalignments with disciplinary facts.
For instance, in a physics reasoning scenario of Fig.~\ref{fig:intro_case}, a generative model tasked with drawing a parallel circuit might mistakenly connect in series.
Identifying such a flaw goes far beyond pixel-level anomaly detection, which requires physical understanding to recognize the structural violation.
Moreover, merely flagging these errors with a binary judgement is insufficient for further correction \cite{chen2026edudiag,chen2024learning}.
To effectively rectify scientific diagrams, a verifier must explicitly articulate \textit{why} the image is incorrect (\eg, pointing out the erroneous series connection) and prescribe exactly \textit{how} to fix it (\eg, providing specific editing instructions).
Thus, we refer to this triplet (\ie, binary judgement, explanation, and actionable editing instruction) as our explainable scientific image verification, which is overlooked by current multimodal verifiers.

Despite recent progress in multimodal verification, two major limitations hinder its application to scientific image generation.
\textbf{First}, prior verification benchmarks and reward models \cite{imagereward, rationalrewards} are built for natural images, focusing on generic image-text alignment and human aesthetic preference. They neglect the scientific structural fidelity and domain-specific reasoning required for scientific image generation.
\textbf{Second}, although recent benchmarks for scientific image generation \cite{genexam, proimagebench} introduce fine-grained evaluation rubrics, they serve merely as static and offline testbeds.
Crucially, it still lacks a dedicated verifier that can actively generate explainable and real-time feedback for scientific error correction, providing a practical application for iterative image rectification.

To bridge this gap, we aim to build a benchmark and verifier model for explainable verification of scientific image generation.
\textbf{On the benchmark side}, we construct \textit{SciGen-Verify}, the first benchmark for this verification task, spanning three core domains (\ie, instruction following, multidisciplinary reasoning, and world knowledge) with a hierarchical protocol over the binary judgement, explanation, and editing instruction. To ensure data quality, the benchmark is meticulously constructed through an AI-assisted annotation pipeline coupled with rigorous human expert inspection.
\textbf{On the model side}, we develop \textit{SciGen-Verifier}, an 8B multimodal verifier trained with cold-start supervised fine-tuning followed by a curriculum-based two-stage reinforcement learning pipeline. It is motivated by the observation that SFT failures predominantly stem from incomplete evidence exploration. The first stage RL employs a rubric-guided process reward to enforce comprehensive reasoning coverage, while the second stage uses an outcome reward to align the judgement, explanation and editing instruction with ground-truth annotation.

Our contributions are summarized as follows:
\textbf{(i)} We construct \textit{SciGen-Verify}, the first benchmark dedicated to the explainable verification of scientific image generation with a hierarchical evaluation protocol.
\textbf{(ii)} We train SciGen-Verifier-8B via a curriculum-based two-stage reinforcement learning pipeline to strengthen scientific reasoning exploration and align corrective annotation.
\textbf{(iii)} Empirical results demonstrate that our SciGen-Verifier matches much larger proprietary baselines, and serves as an online critic to progressively rectify errors during iterative test-time scaling.

\section{Related Work}
\label{sec:related_work}
\noindent
\textbf{Multimodal Verification.}
Recent advances in multimodal large language models (MLLMs) stem from evolving reasoning paradigms \cite{visual-rft,llm-i} and unified generative capabilities \cite{blip3-o,onecat}.
To evaluate multimodal understanding,
\citet{vlrewardbench} propose a testbed for multimodal preference alignment. VisualPRM \cite{visualprm} extends to step-wise process verification on reasoning trace.
Meanwhile, the emergence of generative models has spurred efforts to develop automatic verification for text-to-image tasks.
To overcome the coarse semantic similarity of early metrics like CLIPScore \cite{clipscore}, ProImage-Bench \cite{proimagebench} probes fine-grained semantic alignment via question-answer pairs.
Beyond static scoring, explainable feedback generation has been explored for multimodal reasoning errors, from distractor correction in visual commonsense reasoning \cite{chen2024learning} to educational diagnostic reasoning with error tracing and correction \cite{chen2026edudiag}, whereas scientific image generation still lacks a dedicated verifier that articulates why an image fails and how to fix it.

\noindent
\textbf{Scientific Image Generation.}
Most generative models are confined to open-domain natural scenes and struggle to satisfy scientific image generation demands.
To bridge this gap, emerging works explore domain-specific generation via code rendering \cite{vthinker,tikzero} or direct visual synthesis \cite{autofigure}, and SciIR \cite{ma2026sciir} further provides a large-scale training dataset and benchmark for scientific image generation.
For evaluation, some multidisciplinary benchmarks (\eg, MMMG \cite{mmmg}, SridBench \cite{sridbench} and ScImage \cite{scimage}) rely on holistic metrics.
While recent frameworks (\eg, GenExam \cite{genexam}, ProImage-Bench \cite{proimagebench}) advance to structural rubrics for fine-grained evaluation, they merely serve as static testbeds without explainable error correction.

\section{\textit{SciGen-Verify} Benchmark}
\label{sec:dataset}
\subsection{Benchmark Overview}
To systematically assess the verification of scientific visual outcomes, we introduce \textit{SciGen-Verify}, a comprehensive benchmark with three core domains: \textbf{(i)} Instruction Following tests the MLLMs' adherence to multi-step constraints for synthesizing or editing scientific images; \textbf{(ii)} Multidisciplinary Reasoning probes logical and structural fidelity in complex STEM diagrams (\eg, figures strictly governed by physical or geometric laws); and \textbf{(iii)} World Knowledge assesses reliance on domain-specific factual fidelity.
Moving beyond generic holistic evaluation, \textit{SciGen-Verify} formulates this verification as an executable reasoning task.
Specifically, each sample is annotated with a binary correctness judgement, with a detailed explanation. Crucially, for anomalous outputs (\ie, false judgement), the MLLM is required to generate actionable editing instructions for error correction, offering a pathway for image rectification.
In the following, we describe the construction process of \textit{SciGen-Verify}, as shown in Fig.~\ref{pic_dataset_cons}.
More details are provided in Appendix~\ref{app:scigen_verify}.
\begin{figure}[!t]
  \centering
  \includegraphics[width=\textwidth]{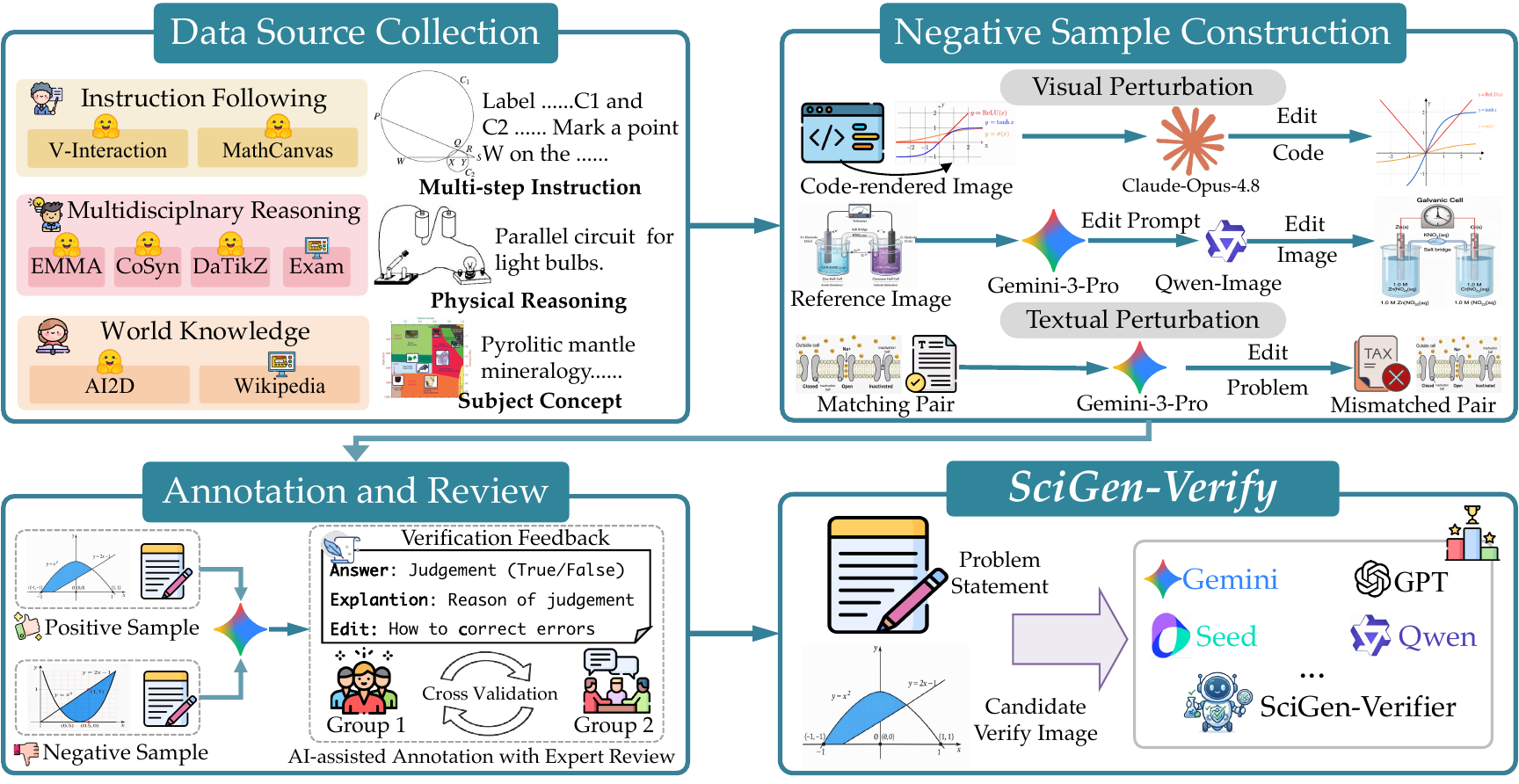}
 \caption{Construction workflow of our \textit{SciGen-Verify} benchmark, including data collection, negative sample construction, and annotation with expert review.}
\label{pic_dataset_cons}
\end{figure}

\subsection{Data Source}
To ensure our benchmark covers real-world scientific image generation scenarios, we collect high-quality data samples spanning three core domains: Instruction Following, Multidisciplinary Reasoning, and World Knowledge.
Since these sources vary in format and annotation granularity, we perform different preprocessing methods to standardize them for further unified verification.

\vspace{0.3em}
\noindent\textbf{Instruction Following.}
This domain requires the verifier to accurately assess adherence to complex, multi-step textual constraints for diagram synthesis and modification (\eg, \textit{``Draw a coordinate plane, plot the points A(1, 2) and B(4, 2), and add the line segment AB with its midpoint labeled M\dots''}). We collect data from V-Interaction \cite{vthinker} and MathCanvas \cite{mathcanvas} datasets.
Specifically,
MathCanvas directly provides images and potential textual instruction pairs for image generation and editing tasks.
V-Interaction consists of interleaved reasoning trajectories where intermediate visual images are updated via code.
While it provides the ``before'' and ``after'' code-rendered images for each reasoning step, it lacks explicit natural language editing instruction.
Hence, we prompt Claude-Sonnet-5 \cite{claude4.6sonnet} to analyze the code differences between adjacent visual images and generate the corresponding natural language editing instructions.

\vspace{0.3em}
\noindent\textbf{Multidisciplinary Reasoning.}
Beyond explicit instruction following, this domain evaluates whether verifiers can assess scientific diagrams that serve as answers to discipline-specific reasoning problems.
It requires verifier to interpret subject-specific problems and infer the expected diagrammatic solution, such as circuit connections, chemical structures, function plots.
To construct this subset, we aggregate data from three primary streams.
First, we collect scientific diagrams from CoSyn \cite{CoSyn} and DaTikZ-V3 \cite{datikz-v3}, which contain code-rendered figures across mathematics, physics, chemistry, and other STEM subjects.
For each sample, we prompt Claude-Opus-4.8 \cite{claude4.6opus} to inspect the corresponding code and synthesize reasoning-intensive problem statements, for which the original figure serves as the ground-truth answer.
Second, we adapt multiple-choice questions from EMMA \cite{emma} by selecting samples whose options are unique images.
We then reformulate them into scientific image generation tasks, where correct options serve as reference images.
Finally, we crawl educational examination papers and extract authentic diagram-construction problems, \eg, drawing a circuit according to component constraints.

\vspace{0.3em}
\noindent\textbf{World Knowledge.}
Beyond explicit instruction following and multidisciplinary reasoning, this domain evaluates whether the verifier ensures the visual content faithfully aligns with established scientific concepts (\eg, illustrating the stratigraphic layers of the Earth's mantle).
First, we systematically mine Wikipedia with curated discipline-specific taxonomy tags and keywords. Then, we crawl a wide spectrum of professional scientific images.
Second, we choose images from AI2D \cite{ai2d}, which features diverse pedagogical visualizations, including ecological food webs, microscopic cellular anatomies, and macroscopic geophysical processes.
Since both sources provide visual assets without textual descriptions, we employ Gemini-3-Flash \cite{gemini3} to generate concept-centric captions capturing scientific entities and relations as problem statements.

\subsection{Benchmark Curation Pipeline}
We construct \textit{SciGen-Verify} through a systematical curation pipeline to ensure both comprehensive disciplinary coverage and annotation reliability.
Based on the collected data sources,
we sample a subset according to their disciplinary taxonomies.
We construct corresponding negative samples via controlled perturbations.
To support explainable verification, we finally utilize AI-assisted annotation followed by rigorous expert review to equip these samples with reliable judgement and detailed explanation and potential editing instruction.

\subsubsection{Negative Sample Construction}
To evaluate the verifier's robustness against genuine scientific mistakes rather than trivial mismatches, we must ensure the constructed negative samples are highly challenging and realistic. Inspired by \cite{omniverifier}, we design the following two automated strategies to construct negative samples.

\vspace{0.3em}
\noindent\textbf{Visual Perturbation.}
We construct negative data by modifying the visual output while keeping the original problem statements fixed.
Depending on the image format, we process them as follows:
For samples equipped with rendering code, we prompt Claude-Sonnet-5 to identify error-prone elements within the scientific context and subtly modify the code to render an incorrect diagram (\eg, bypassing a resistor).
For other reference images, inspired by \cite{supercorrct}, we employ Gemini-3-Pro to simulate an experienced teacher. By analyzing the input problem statement and reference image, the model identifies tricky knowledge points and common student misconceptions, subsequently outputting a corresponding editing instruction. We then feed this instruction and the original image into Qwen-Image-Edit \cite{qwen-image} to generate a rigorously flawed candidate image.
Furthermore, to mitigate unintended artifacts introduced by the generative model, we introduce an automated filtering stage with Gemini-3-Flash. Before any human review, it scans the generated candidate images to discard low-quality samples—specifically targeting severe blurriness, illegible text, or clear failures against the editing instruction.
This ensures the visual errors reflect authentic human student cognitive mistakes rather than poor generation fidelity.

\vspace{0.3em}
\noindent\textbf{Textual Perturbation.}
Alternatively, we construct negative data by altering the textual problem statements while retaining the original correct candidate image.
We utilize Gemini-3-Pro to deeply inspect the critical solving steps and generative requirements of the original problem.
Specifically, the model first extracts key scientific entities, quantitative parameters, and structural relationships from the prompt. It then systematically injects counterfactual conditions by altering specific details, such as reversing structural logic (\eg, changing a required ``parallel'' circuit connection to ``series''), modifying numerical values (\eg, altering a specified geometric angle), or substituting domain-specific attributes (\eg, replacing a required chemical functional group).
This strategy creates a logical counterfactual mismatch between the text and candidate image.

\subsubsection{Annotation and Expert Review}
\vspace{0.3em}
\noindent\textbf{AI-Assisted Initial Annotation.}
We first utilize Gemini-3.1-Pro to generate comprehensive initial annotations.
Specifically, for the binary judgement, the model is prompted to output a step-by-step reasoning rationale as explanation. For negative samples, it also synthesizes a precise editing instruction designed to rectify the identified visual errors in the false image.

\vspace{0.3em}
\noindent\textbf{Expert Review and Refinement.}
To ensure the benchmark's rigor, we recruit 10 domain experts---comprising two graduate students for each of the five target disciplines (Mathematics, Physics, Chemistry, Biology, and Geography)---to conduct a comprehensive human evaluation.
To prevent prior cognitive bias, paired positive and negative samples are partitioned into two separate groups. Each sample is independently reviewed by domain-matched experts, and it is retained only if both experts explicitly approve its correctness and necessary refinements.
The process is structured around two core stages:
\textbf{(i) Multi-Dimensional Quality Filter.} The experts evaluate the samples based on three strict criteria: \textit{Visual Fidelity}, filtering out generated images with severe blurriness or rendering artifacts; \textit{Task Difficulty}, discarding overly simplistic cases that lack reasoning depth; and \textit{Annotation Correctness}, verifying the strict factual accuracy of the binary judgements, reasoning rationales, and editing instructions.
\textbf{(ii) Manual Refinement and Verification.} Samples with unresolvable ambiguities or fatal visual errors are directly discarded. For minor textual flaws (\eg, slight hallucinations or vague instructions), the first expert from the corresponding discipline manually rewrites the text to ensure the rationales are logical and the editing instructions are actionable. Subsequently, the second expert independently reviews these modifications.
The sample is retained only if the second expert approves the refined results.
Ultimately, this rigorous process yields the finalized \textit{SciGen-Verify} benchmark, comprising 1,350 data samples: 406 for instruction following, 287 for multidisciplinary reasoning, and 657 for world knowledge.

\subsection{Evaluation Methods}
We evaluate verifiers using a strict three-tier hierarchical protocol. Detailed formulations are provided in Appendix \ref{sec:appendix_eval}.
We design three cascading metrics:  judgement ($\text{Acc}_a$), explanation ($\text{Acc}_e$), and editing instruction Accuracy ($\text{Acc}_i$).
Specifically, $\text{Acc}_a$ is a rule-based metric that solely measures binary prediction correctness, whereas $\text{Acc}_e$ and $\text{Acc}_i$ are model-based metrics evaluated by proprietary LLMs to assess logical consistency and actionable correction. Both judges are validated against three human experts on 300 sampled instances, where the judge--human agreement ($\kappa \geq 0.76$) approaches the human--human ceiling ($\kappa = 0.83$), and the evaluation is further shown to be robust to the choice of judge (Appendix~\ref{app:judge_reliability}).

\section{SciGen-Verifier Model}
\label{sec:model}
The inherent complexity of scientific image verification motivates us to investigate methods that can progressively strengthen the capabilities of current advanced MLLMs.
We therefore develop the reasoning-augmented SciGen-Verifier through a two-stage training pipeline: supervised fine-tuning (SFT) cold start (Sec. \ref{sec:sft}) that endows the model with structured reasoning output and basic verification ability, followed by curriculum-based reinforcement learning (RL) phase (Sec. \ref{sec:rl}) with progressive reward signals to further strengthen its reasoning depth and output reliability. The overall pipeline is illustrated in Fig.~\ref{fig:method_overview} (Appendix~\ref{app:verifier_training}).

\subsection{Supervised Fine-Tuning}
\label{sec:sft}

\paragraph{SFT Data Construction.}
The SFT corpus is drawn from the same upstream sources as \textit{SciGen-Verify} (Sec.~\ref{sec:dataset}), and the two are split at the source-sample level so that a source problem or image, together with all of its perturbed variants, belongs to exactly one set, ensuring strictly no data overlap (see Appendix~\ref{app:verifier_training}). To efficiently scale annotations, we utilize Seed-2.0-Pro \cite{seed2.0} to generate structured responses strictly following an XML-style template: ``\texttt{<think>} reasoning trace \texttt{</think>} \texttt{<answer>} binary judgement \texttt{</answer>} \texttt{<explanation>} rationale \texttt{</explanation>} \texttt{<edit>} editing instruction for false instances \texttt{</edit>}''.
To guarantee quality, we apply rejection sampling \cite{seed-1.5-vl} to retain only instances where the binary judgement matches the ground truth and the explanation is validated by Gemini-3-Flash.
It finally yields 48,311 training instances.

\paragraph{Training.}
We instantiate SciGen-Verifier from Qwen3-VL-8B-Instruct \cite{qwen3-vl} and perform full-parameter supervised fine-tuning on the curated corpus $\mathcal{D}_{\text{SFT}}$. We optimize the model $\pi_\theta$ via the standard next-token prediction objective:
\begin{equation}
\mathcal{L}_{\text{SFT}}(\theta) = -\mathbb{E}_{(x,y) \sim \mathcal{D}_{\text{SFT}}} \left[ \log \pi_\theta(y \mid x) \right],
\label{eq:sft-loss}
\end{equation}
where $x$ denotes the problem statement with potential input images and $y$ denotes the curated tagged response.
This stage transfers structured reasoning and subject-specific verification knowledge into the verifier, which is a reliable initialization for subsequent reinforcement learning.

\subsection{Reinforcement Learning}
\label{sec:rl}
After SFT, SciGen-Verifier acquires basic verification ability and outputs with structured format.
However, its judgement remains unreliable on hard instances since the reasoning trace often lacks subject-specific evidence.
Toward this end, we adopt GRPO \cite{grpo} to perform a curriculum-based reinforcement learning pipeline: the first stage sharpens the verifier's judgement through a rubric-guided reward, while the second stage directly refines its judgement, explanations, and editing instructions through an outcome reward.
We use the SFT model to perform $K=8$ rollout sampling across the SFT corpus and unseen complementary data.
Inspired by \cite{polaris}, we evaluate the binary judgements using rule-based metrics and retain 15\textit{K} queries that yield partial success (\ie, $k \in \{1, \dots, 7\}$ correct rollouts) to provide appropriate challenging data for RL.

\subsubsection{Motivation.}
Error analysis reveals that SFT failures primarily stem from incomplete evidence exploration rather than incoherent reasoning. To address this, we design a two-stage RL curriculum.
Since valid explanations and editing instructions strictly depend on correct initial judgements, applying full outcome-level rewards too early would lead to unstable training.
The first stage employs a rubric-guided and process-level reward, enforcing comprehensive coverage of instance-specific checkpoints in reasoning traces for reliable judgement.
Next, the second stage introduces an outcome reward only after robust reasoning is established, explicitly aligning the final judgement, explanation and editing instruction with ground-truth annotations.

\begin{table*}[t]
\centering
\caption{\label{tab:main_res} Evaluation results of different MLLMs on the \textit{SciGen-Verify}. The best and second-best performance in each column are highlighted in \textbf{bold} and \underline{underlined}, respectively. The overall denotes the macro-average accuracy.}
\setlength{\tabcolsep}{3pt}
\renewcommand{\arraystretch}{1.25}
\resizebox{\textwidth}{!}{
\begin{tabular}{l|ccc|ccc|ccc|ccc}
\toprule[1pt]
\multirow{2}{*}{\textbf{Models}} & \multicolumn{3}{c|}{\textbf{Instruction}} & \multicolumn{3}{c|}{\textbf{Reasoning}} & \multicolumn{3}{c|}{\textbf{Knowledge}} & \multicolumn{3}{c|}{\textbf{Overall}} \\ \cmidrule{2-13}
                        & Acc$_a$       & Acc$_e$     & Acc$_i$      & Acc$_a$   & Acc$_e$  & Acc$_i$  & Acc$_a$   & Acc$_e$ & Acc$_i$ & Acc$_a$  & Acc$_e$ & Acc$_i$ \\ \midrule[1pt]
GPT-6-Astra               & \underline{78.16} & 71.84          & \underline{58.74} & 78.65          & 76.78          & 59.55          & 70.33          & 69.86          & 42.48          & 75.71          & 72.83          & 53.59 \\
Gemini-3-Flash            & 72.66          & 66.50          & 52.30          & 87.46          & 81.47          & 65.06          & 84.47          & 82.65          & 53.73          & 81.53          & 76.87          & 57.03 \\
Gemini-3.1-Flash-Lite     & 63.55          & 54.19          & 41.39          & 77.35          & 68.64          & 50.79          & 78.39          & 72.60          & 34.31          & 73.10          & 65.14          & 42.16 \\
Gemini-3.1-Pro            & 75.04          & 68.45          & 51.21          & 85.93          & 79.94          & 55.39          & 84.62          & 82.93          & 57.98          & 81.86          & 77.11          & 54.86 \\
Gemini-3.5-Flash          & 72.12          & 69.57          & 51.69          & 86.55          & \underline{84.00} & \underline{68.12} & 85.47          & 82.69          & 49.80          & 81.38          & 78.75          & 56.54 \\
Seed-1.8                  & 56.90          & 48.28          & 32.23          & 86.76          & 72.47          & 58.18          & 81.13          & 75.91          & 42.45          & 74.93          & 65.55          & 44.28 \\
Seed-2.0-Pro              & 69.21          & 59.85          & 48.65          & 85.02          & 77.19          & 61.49          & 84.17          & 80.49          & 51.71          & 79.47          & 72.51          & 53.95 \\
MiMo-V2-Omni              & 63.04          & 53.75          & 36.90          & 82.63          & 71.86          & 51.00          & 77.56          & 70.60          & 36.84          & 74.41          & 65.40          & 41.58 \\
Qwen3-VL-8B-Thinking      & 51.36          & 36.30          & 20.52          & 74.47          & 57.91          & 32.60          & 57.34          & 45.33          & 17.45          & 61.06          & 46.51          & 23.52 \\
Qwen3-VL-235B-Thinking    & 54.19          & 42.47          & 29.43          & 78.67          & 62.46          & 43.55          & 69.71          & 63.25          & 29.28          & 67.52          & 56.06          & 34.08 \\
Qwen3.5-397B              & 67.65          & 59.26          & 50.94          & 85.37          & 77.97          & 58.66          & 69.97          & 64.27          & 37.67          & 74.33          & 67.17          & 49.09 \\
Qwen3.8-Max               & \textbf{79.33} & \textbf{76.74} & \textbf{60.32} & \textbf{91.01} & \textbf{89.51} & \textbf{74.13} & \underline{87.76} & \underline{87.12} & \underline{59.81} & \underline{86.03} & \textbf{84.46} & \textbf{64.75} \\
GLM-5.3-Flash             & 75.66          & \underline{72.75} & 55.97          & 85.96          & 83.33          & 59.17          & 85.14          & 82.27          & 49.11          & 82.25          & 79.45          & 54.75 \\
DeepSeek-V4-Flash & 67.00          & 59.90          & 42.66          & 79.79          & 76.00          & 55.17          & 72.91          & 71.72          & 40.68          & 73.23          & 69.21          & 46.17 \\
OmniVerifier-7B           & 45.32          & 35.96          & 6.02           & 61.32          & 42.51          & 13.86          & 69.56          & 59.51          & 8.30           & 58.73          & 45.99          & 9.39 \\ \midrule[1pt]
SciGen-Verifier-8B        & 75.62          & 70.38          & 48.93          & \underline{89.02} & 83.47          & 57.84          & \textbf{94.21} & \textbf{91.53} & \textbf{65.48} & \textbf{86.28} & \underline{81.79} & \underline{57.42} \\ \midrule[1pt]
\end{tabular}}

\end{table*}

\subsubsection{RL with Rubric-guided Reward}

\paragraph{Rewarding CoT with Rubrics.}
To supervise the reasoning process rather than solely the final judgement, our first-stage reward evaluates whether the generated reasoning trace explicitly mentions checkpoints defined in an instance-specific scientific rubric (\eg, ``Is the curve passing through the origin?''). These rubrics are produced by a dedicated rubric generator, which is trained on data from the same upstream sources.
Following \cite{rubricbench}, we also rigorously validate the rubric generator on test data (Table~\ref{tab:rubric_performance}), proving its reliability (see Appendix~\ref{app:rubric_performance}).
For each rollout $\tau$ produced by the verifier policy $\pi_\theta$, a coverage checker $\pi_\phi$ assesses whether each checkpoint $r_j$ is correctly addressed within $\tau$'s reasoning trace, yielding a binary indicator $c_j(\tau) \in \{0, 1\}$. The rubric-guided reward is then defined as the coverage ratio over the checkpoints within rubrics:
\begin{equation}
\mathcal{R}_{\text{rubric}}(\tau) = \frac{1}{n} \sum_{j=1}^{n} c_j(\tau).
\label{eq:rubric-reward}
\end{equation}

\paragraph{Total Reward.}
We simultaneously introduce a binary judgement reward $\mathcal{R}_{\text{judge}}$.
To balance process-level exploration and correctness optimization, the first-stage reward combines the judgement and rubric signals via dynamic coefficients:
\begin{equation}
\mathcal{R}_{1}(\tau) = \alpha_t \mathcal{R}_{\text{judge}}(\tau) + \beta_t \mathcal{R}_{\text{rubric}}(\tau).
\label{eq:total-reward}
\end{equation}

Crucially, rather than employing fixed weights, $\alpha_t$ and $\beta_t$ follow a curriculum-like schedule driven by the policy's training mastery (\ie, moving average of judgement accuracy). In early training phases when accuracy is low, the rubric weight $\beta_t$ is kept relatively high to encourage comprehensive exploration of visual evidence and scientific constraints. As the verifier's accuracy improves, the weighting progressively shifts to dominate $\alpha_t$, transitioning the focus toward strict judgement correctness. This adaptive strategy ensures that the auxiliary rubric signal guides reasoning without overshadowing the primary objective of accurate decision-making.
Appendix~\ref{appendix:dynamic-weighting} shows the detailed formulations.

\subsubsection{RL with Outcome Reward}
The second stage directly rewards the textual outputs. Let $e$ and $u$ denote the generated explanation and editing instruction, with ground truths $e^*$ and $u^*$. We define $\mathcal{R}_{\text{exp-edit}}(\tau) = \mathbb{I}(e \cong e^*) + \mathbb{I}(u \cong u^*)$, where $\cong$ denotes semantic equivalence judged by an LLM: GLM-4.7 \cite{glm4.7} for explanations and Doubao-Seed-2.0-Lite \cite{seed2.0} for image-conditioned edits (prompts in Appendix~\ref{app:judge_prompts}). Adding the binary judgement reward yields the stage-2 reward:
\begin{equation}
\mathcal{R}_{2}(\tau) = \mathcal{R}_{\text{judge}}(\tau) + \mathcal{R}_{\text{exp-edit}}(\tau).
\label{eq:stage2-total-reward}
\end{equation}

\section{Experiments}
\label{sec:experiment}

\begin{table*}[t]
\centering
\caption{\label{tab:rl_strategies} Evaluation on training strategies. $\mathcal{S}$ denotes SFT. $\mathcal{R}_1$ and $\mathcal{R}_2$ denote the stage 1 and 2 RL.}
\setlength{\tabcolsep}{5pt}
\renewcommand{\arraystretch}{1.}
\resizebox{\textwidth}{!}{
\begin{tabular}{ccc|ccc|ccc|ccc|ccc}
\toprule[1pt]
\multirow{2}{*}{$\mathcal{S}$} & \multirow{2}{*}{$\mathcal{R}_1$} & \multirow{2}{*}{$\mathcal{R}_2$} & \multicolumn{3}{c|}{\textbf{Instruction}} & \multicolumn{3}{c|}{\textbf{Reasoning}} & \multicolumn{3}{c|}{\textbf{Knowledge}} & \multicolumn{3}{c|}{\textbf{Overall}} \\ \cmidrule{4-15}
 & & & Acc$_a$ & Acc$_e$ & Acc$_i$ & Acc$_a$ & Acc$_e$ & Acc$_i$ & Acc$_a$ & Acc$_e$ & Acc$_i$ & Acc$_a$ & Acc$_e$ & Acc$_i$ \\ \midrule[1pt]
\ding{51} & \ding{55} & \ding{55} & 65.35 & 60.63 & 35.34 & 82.25 & 75.36 & 43.45 & 92.66 & 89.97 & 61.36 & 80.09 & 75.32 & 46.72 \\
\ding{51} & \ding{51} & \ding{55} & 71.18 & 63.72 & 40.35 & 86.06 & 77.35 & 47.94 & 93.61 & 90.42 & 63.11 & 83.62 & 77.16 & 50.47 \\
\ding{51} & \ding{55} & \ding{51} & 69.86 & 65.23 & 42.78 & 84.93 & 79.14 & 51.22 & 93.24 & 91.02 & 64.35 & 82.68 & 78.46 & 52.78 \\
\ding{51} & \ding{51} & \ding{51} & \textbf{75.62} & \textbf{70.38} & \textbf{48.93} & \textbf{89.02} & \textbf{83.47} & \textbf{57.84} & \textbf{94.21} & \textbf{91.53} & \textbf{65.48} & \textbf{86.28} & \textbf{81.79} & \textbf{57.42} \\ \bottomrule[1pt]
\end{tabular}}
\end{table*}

\subsection{Performance Comparison}
Table~\ref{tab:main_res} reports the evaluation results on \textit{SciGen-Verify} benchmark. We have the following findings:
\textbf{(i) Frontier MLLMs still exhibit limited scientific image verification capability.}
Despite their established multimodal reasoning prowess, leading proprietary MLLMs remain far from reliable on \textit{SciGen-Verify}. Even the strongest baseline Qwen3.8-Max (86.03\% overall Acc$_a$) and frontier flagships such as Gemini-3.5-Flash and GPT-6-Astra suffer a precipitous drop once the protocol demands explainable feedback: GPT-6-Astra falls from 75.71\% overall Acc$_a$ to 53.59\% Acc$_i$, and Gemini-3.5-Flash from 81.38\% to 56.54\%, revealing a profound deficiency in fine-grained mistake localization and correction.
More notably, OmniVerifier-7B, a model trained for evaluating generic images, performs poorly on \textit{SciGen-Verify}.
This stark contrast demonstrates that scientific verification extends far beyond generic text-image alignment and demands intricate disciplinary knowledge.
\textbf{(ii) SciGen-Verifier achieves robust verification performance.}
Built upon a lightweight 8B backbone, SciGen-Verifier demonstrates robust generalization across all three core evaluation domains. It achieves the highest overall Acc$_a$ of 86.28\%, outperforming the strongest baseline Qwen3.8-Max (86.03\%) as well as frontier proprietary models such as Gemini-3.5-Flash (+4.90\%) and GPT-6-Astra (+10.57\%). Notably, our model exhibits exceptional proficiency in world knowledge verification, establishing a new state-of-the-art with 94.21\% Acc$_a$, 91.53\% Acc$_e$, and 65.48\% Acc$_i$, which eclipses Qwen3.8-Max by +6.45\%, +4.41\%, and +5.67\%, respectively. These findings demonstrate that our curriculum-based reinforcement learning effectively endows the SFT model with profound scientific deliberation capabilities.
\textbf{(iii) Correcting scientific errors is significantly harder than identifying them.}
As observed in Table~\ref{tab:main_res}, there is a sharp performance drop for all models on the Acc$_i$ metric. While current MLLMs can successfully diagnose scientific inconsistencies, they struggle to prescribe precise modifications. For instance, Gemini-3.5-Flash and GPT-6-Astra are capped at 56.54\% and 53.59\% overall Acc$_i$, respectively, and even the strongest baseline Qwen3.8-Max attains only 64.75\%. Similarly, despite its strong diagnostic capability, our SciGen-Verifier-8B obtains 57.42\%. This performance gap highlights that synthesizing actionable corrective edits remains a major bottleneck in scientific image verification. Overcoming this limitation is crucial to rectify flawed scientific images generated by generative models.

\needspace{7\baselineskip}
\subsection{Effect of Explicit Reasoning}

\begin{wrapfigure}{r}{0.42\textwidth}
\vspace{-8pt}
\centering
\includegraphics[width=\linewidth]{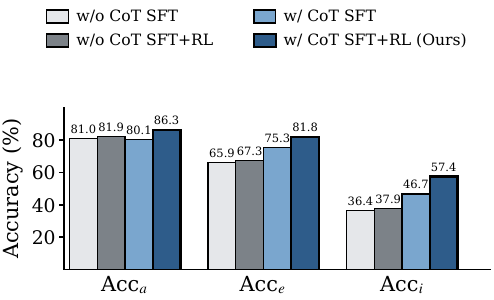}
\vspace{-6pt}
\caption{Effect of CoT reasoning.}
\label{fig:ablation_cot}
\vspace{-8pt}
\end{wrapfigure}
To evaluate the impact of explicit reasoning, we compare models trained with and without chain-of-thought (CoT) across both SFT and RL stages (Fig.~\ref{fig:ablation_cot}). Directly predicting outcomes (w/o CoT SFT) yields a competitive Acc$_a$ of 81.0\% yet suffers a severe performance collapse when articulating faithful feedback (Acc$_e$: 65.9\%; Acc$_i$: 36.4\%). Applying RL to this direct-prediction paradigm (w/o CoT SFT+RL)\footnote{The prompt requires the model to emit the fixed format without \texttt{<think>} field and rubric-guided reward.} brings negligible gains, revealing that RL fundamentally struggles to optimize complex visual editing without intermediate logical grounding. Conversely, integrating CoT (w/ CoT SFT) effectively mitigates this degradation, lifting Acc$_e$ and Acc$_i$ to 75.3\% and 46.7\% with a marginal Acc$_a$ concession. Most crucially, the synergy between CoT and RL (w/ CoT SFT+RL) maximizes the model's potential, propelling overall Acc$_a$ to 86.3\% and Acc$_i$ to 57.4\%, with the most pronounced gains observed in the instruction-following domain. These results firmly establish that intermediate CoT steps are not merely optional features, but essential structural prerequisites enabling RL to synthesize reliable corrections in scientific verification.

\subsection{Effect of Curriculum-based RL Training}
To validate the efficacy of our two-stage curriculum RL, we analyze the performance of SciGen-Verifier across different training strategies (Table~\ref{tab:rl_strategies}). While supervised fine-tuning ($\mathcal{S}$) establishes foundational capabilities, directly applying outcome-level rewards without process-level grounding ($\mathcal{S} + \mathcal{R}_2$) yields suboptimal improvements, prematurely capping reasoning Acc$_a$ at 84.9\% and Acc$_i$ at 51.2\%. This highlights a critical risk of premature convergence in RL: without explicitly encouraging initial exploration within reasoning traces, the model exploits shortcuts and settles into local optima. Our decoupled curriculum effectively circumvents this limitation. Stage 1 RL, governed by rubric rewards, enforces rigorous visual evidence exploration, pushing reasoning Acc$_a$ to 86.1\%. Building upon this expanded reasoning capacity, Stage 2 introduces outcome-level alignment to teach precise articulation. This synergy successfully unlocks the model's true upper bound, propelling overall Acc$_a$ to 86.3\% and Acc$_i$ to 57.4\%, confirming that initial rubric-guided exploration is an indispensable prerequisite for maximizing the potential of scientific image verification.

\begin{figure*}[t]
  \centering
  \includegraphics[width=\textwidth]{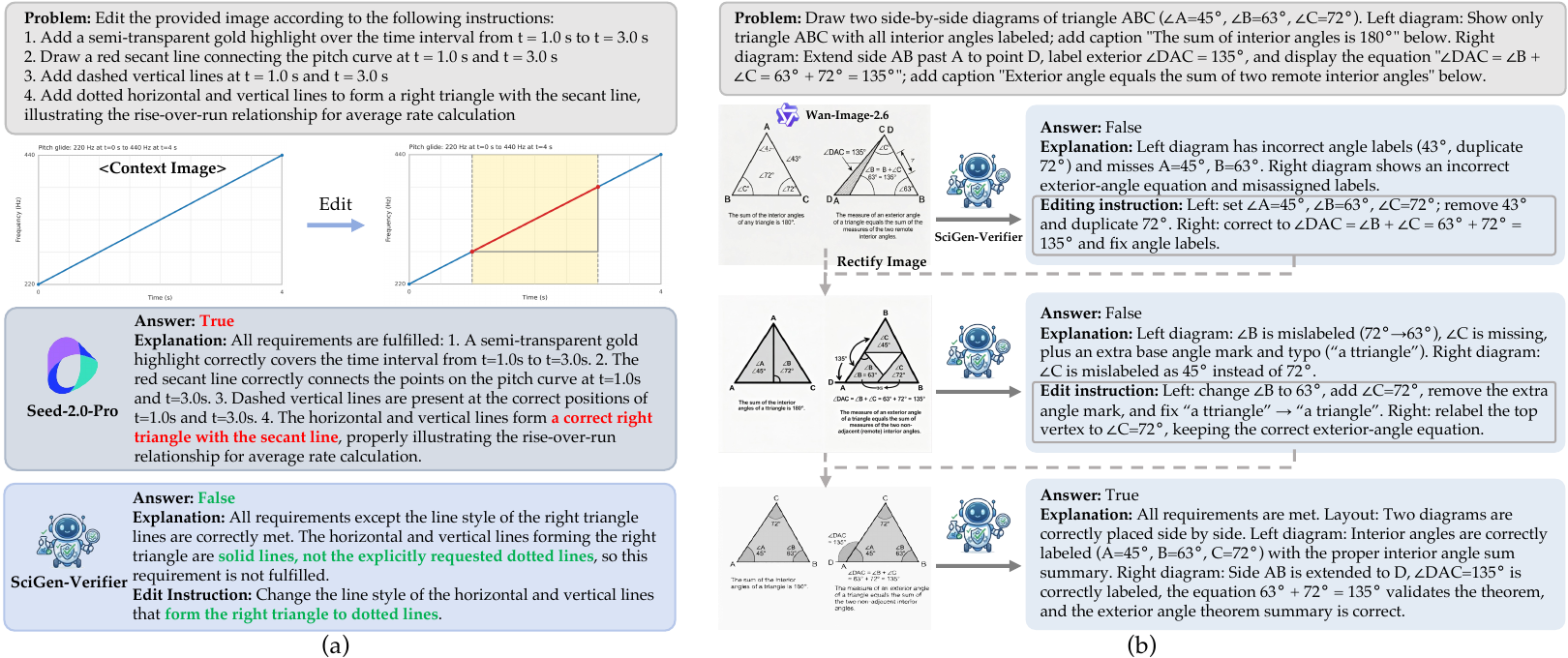}
  \caption{Case study. (a) Comparison with a strong baseline. (b) Iterative test-time scaling for image rectification with SciGen-Verifier as the online critic.}
  \label{fig:case_study}
\end{figure*}

\subsection{Qualitative Analysis}
As shown in Fig.~\ref{fig:case_study} (a), while Seed-2.0-Pro exhibits a false-positive hallucination by overlooking the explicitly requested dotted lines, our model precisely localizes this subtle visual violation and formulates actionable editing instructions.
Beyond benchmark performance, Fig.~\ref{fig:case_study} (b) illustrates the deployment of SciGen-Verifier as an online critic for image rectification with iterative test-time scaling (TTS) (Appendix~\ref{app:tts}).
By iteratively diagnosing geometric mislabeling and mathematical inconsistencies, our model provides precise critiques that guide the generative model to self-correct.

\section{Conclusion}
This paper focuses on explainable verification for scientific image generation.
We first construct \textit{SciGen-Verify} benchmark for this verification task, covering three representative domains.
Second, we develop \textit{SciGen-Verifier} through a curriculum-based two-stage reinforcement learning pipeline, where rubric-guided process rewards strengthen reasoning exploration and outcome rewards explicitly align with ground truth.
Extensive experiments confirm that explicit reasoning and curriculum-based reinforcement learning are jointly indispensable, and show that SciGen-Verifier serves as an online critic for iterative image rectification.

\bibliography{iclr2027_conference}
\bibliographystyle{iclr2027_conference}
\clearpage
\appendix
\section{Details of \textit{SciGen-Verify}}
\label{app:scigen_verify}

\subsection{Annotation Protocol and Human Verification}
\label{sec:appendix_annotation}

Each annotated example in \textit{SciGen-Verify} contains a binary judgement, an explanation, and, when the candidate image is inconsistent with the instruction, an editing instruction. For positive examples, the annotation explains why the candidate image satisfies the scientific image generation or editing instruction by explicitly mapping visual evidence to textual constraints. For negative examples, the annotation identifies the violated scientific facts, structural relations, or instruction-following requirements, and then provides an actionable editing instruction for correcting the image.

\vspace{0.5em}
\noindent\textbf{AI-Assisted Annotation.}
We use MLLM-based annotation to generate initial explanations and editing instructions with the prompt templates in Appendix~\ref{app:annotation_prompts}. The annotation model is instructed to inspect the problem statement and candidate image jointly, and to ground its explanation in concrete visual evidence rather than generic comments. For false examples, the generated editing instruction must describe localized and executable changes that would make the image consistent with the original instruction.

\vspace{0.5em}
\noindent\textbf{Human Verification.}
Following the review protocol in Sec.~\ref{sec:dataset}, every sample is independently reviewed by two domain-matched experts, who check the binary judgement, explanation, and editing instruction. Samples with unresolvable ambiguities, severe rendering failures, or unsupported scientific claims are discarded. For minor textual defects, the first expert rewrites the explanation or editing instruction, and the sample is retained only after the second expert approves the refined result.

\subsection{Evaluation Protocol}
\label{sec:appendix_eval_protocol}

For benchmark inference, all verifiers use the same prompt template shown in Appendix~\ref{app:verifier_prompt}. Given the scientific image generation or editing instruction and the candidate image, the verifier is required to output a binary answer, an explanation, and, when the answer is false, an editing instruction. We then compute the hierarchical metrics in Appendix~\ref{sec:appendix_eval}, where explanation and editing-instruction quality are evaluated with the judge prompts in Appendix~\ref{app:judge_prompts}. This protocol ensures that all baseline models and SciGen-Verifier are evaluated under the same input-output interface.

\subsection{Evaluation Metric}
\label{sec:appendix_eval}
We evaluate verifiers using a strict three-tier hierarchical protocol. Let $N$ be the total number of samples and $N_{\text{neg}}$ be the number of negative samples. For the $i$-th sample, let $y_i, \hat{y}_i \in \{\text{True}, \text{False}\}$ denote the ground-truth and predicted judgements, respectively. The benchmark provides a ground-truth explanation $e_i$ for all samples. The evaluated model is required to generate an explanation $\hat{e}_i$ for every sample, and an editing instruction $\hat{c}_i$ only when it predicts $\hat{y}_i=\text{False}$. These outputs are subsequently assessed by advanced models serving as automated judges. For text-only judge tasks, we use GLM-4.7 with thinking enabled; for judge tasks involving image input, we use Doubao-Seed-2.0-Lite with medium reasoning effort. We also conduct manual spot checks on sampled annotations to ensure that the automated evaluation remains consistent with expert judgement. Let $\mathbb{I}(\cdot)$ be the indicator function. To penalize random guessing, our evaluation cascades: an explanation is scored only if the judgement is correct, and an editing instruction is evaluated only if both the judgement and explanation are correct.

\vspace{0.5em}
\noindent\textbf{Binary Judgement Accuracy ($\text{Acc}_a$).}
This foundational metric evaluates if the model correctly identifies whether the image aligns with the problem statement:
\begin{equation}
\text{Acc}_a = \frac{1}{N} \sum_{i=1}^{N} \mathbb{I}(\hat{y}_i = y_i).
\end{equation}

\vspace{0.5em}
\noindent\textbf{Explanation Accuracy ($\text{Acc}_e$).}
This metric demands logical consistency for all samples. We employ GLM-4.7 \cite{glm4.7} with thinking enabled, denoted as $\mathcal{F}_{\text{exp}}$, to compare $\hat{e}_i$ against $e_i$ conditioned on the problem statement and the ground-truth judgement. A prediction is correct only if the judgement matches and the explanation is valid:
\begin{equation}
\text{Acc}_e = \frac{1}{N} \sum_{i=1}^{N} \Big( \mathbb{I}(\hat{y}_i = y_i) \cdot \mathbb{I}\big(\mathcal{F}_{\text{exp}}(e_i, \hat{e}_i)\big) \Big).
\end{equation}

\vspace{0.5em}
\noindent\textbf{Instruction Accuracy ($\text{Acc}_i$).}
For negative samples, the verifier must provide actionable feedback. We use Doubao-Seed-2.0-Lite \cite{seed2.0} with medium reasoning effort, denoted as $\mathcal{F}_{\text{edit}}$, to evaluate whether the predicted editing instruction $\hat{c}_i$ can correct the flawed image according to the instruction, candidate image, and reference editing instruction. $\text{Acc}_i$ is computed as a negative-sample conditional metric:
\begin{equation}
\begin{aligned}
\text{Acc}_i = &\frac{1}{N_{\text{neg}}} \sum_{i: y_i = \text{False}} \Big( \mathbb{I}(\hat{y}_i = y_i) \cdot \\
& \mathbb{I}\big(\mathcal{F}_{\text{exp}}(e_i, \hat{e}_i)\big) \cdot \mathbb{I}\big(\mathcal{F}_{\text{edit}}(c_i, \hat{c}_i)\big) \Big),
\end{aligned}
\end{equation}
where $N_{\text{neg}}$ denotes the total number of negative samples in the benchmark and $c_i$ is the reference editing instruction. Since $\text{Acc}_i$ is normalized over negative samples only, it should be interpreted as a conditional editing-instruction metric rather than a quantity directly comparable to the full-dataset accuracies $\text{Acc}_a$ and $\text{Acc}_e$. The cascading formulation still ensures that an editing instruction is credited only after both the binary judgement and explanation are judged correct.

\subsection{Reliability of Model-based Judges}
\label{app:judge_reliability}
\noindent\textbf{Agreement with Human Experts.}
To validate the automated judges, we sample 300 instances from \textit{SciGen-Verify} and have them independently labeled by three domain experts; instances with unanimous agreement among the three experts serve as the reference. Table~\ref{tab:judge_agreement} reports the accuracy, F1, and Cohen's $\kappa$ of both judges against this reference, together with the human--human inter-annotator agreement (IAA) as the ceiling. Both judges achieve high $\kappa$ values, and their agreement with human experts is close to the human--human ceiling, validating the reliability of the model-based metrics.

\begin{table}[htbp]
    \centering
    \caption{Agreement between the automated judges and human experts on 300 sampled instances. IAA denotes the human--human inter-annotator agreement.}
    \label{tab:judge_agreement}
    \setlength{\tabcolsep}{6pt}
    \renewcommand{\arraystretch}{1.15}
    \begin{tabular}{lccc}
        \toprule
        \textbf{Agreement} & \textbf{Acc. (\%)} & \textbf{F1 (\%)} & \textbf{Cohen's $\kappa$} \\
        \midrule
        Human--Human (IAA) & 98.3 & 99.2 & 0.83 \\
        $\text{Acc}_e$ judge (GLM-4.7) vs. Human & 95.3 & 97.2 & 0.79 \\
        $\text{Acc}_i$ judge (Doubao-Seed-2.0-Lite) vs. Human & 94.7 & 96.8 & 0.76 \\
        \bottomrule
    \end{tabular}
\end{table}

\noindent\textbf{Robustness to the Judge Choice.}
Since the editing-instruction judge belongs to the Seed family, we further examine whether it favors Seed-family baselines. On the same 300 human-labeled instances, the judge agrees with human experts on 94.3\% of Seed-family outputs and 94.8\% of the outputs from other models; the two numbers are close, indicating no family-related preference. We additionally re-evaluate the overall $\text{Acc}_i$ with GPT-5-mini, a third-party judge outside both our pipeline and the baselines (Table~\ref{tab:judge_swap}). The two judges agree closely on individual decisions, and the model ranking is fully preserved, confirming that our evaluation is robust to the judge choice.

\begin{table}[htbp]
    \centering
    \caption{Overall $\text{Acc}_i$ under the default judge (Doubao-Seed-2.0-Lite) and a third-party judge (GPT-5-mini). $\Delta$ denotes the difference between the two judges.}
    \label{tab:judge_swap}
    \setlength{\tabcolsep}{6pt}
    \renewcommand{\arraystretch}{1.15}
    \begin{tabular}{lccc}
        \toprule
        \textbf{Model} & \textbf{Doubao judge} & \textbf{GPT-5-mini judge} & $\Delta$ \\
        \midrule
        SciGen-Verifier-8B & 57.42 & 57.07 & $-$0.35 \\
        Gemini-3.1-Pro & 54.86 & 55.21 & $+$0.35 \\
        Seed-2.0-Pro & 53.95 & 53.30 & $-$0.65 \\
        Seed-1.8 & 44.28 & 43.98 & $-$0.30 \\
        MiMo-V2-Omni & 41.58 & 41.02 & $-$0.56 \\
        \bottomrule
    \end{tabular}
\end{table}

\section{Details of \textit{SciGen-Rubric}} \label{app:scigen-rubric}
\subsection{Dataset Construction}
\label{sec:scigen_rubric_bench}

To enable fine-grained, rubric-based evaluation and alignment for scientific figure generation, we construct \textit{SciGen-Rubric}, a hierarchical rubric dataset spanning the same three capability dimensions as \textit{SciGen-Verify}: Instruction Following, Multidisciplinary Reasoning, and World Knowledge.
The raw data is drawn from the data sources described in Section~\ref{sec:dataset}.
The resulting dataset comprises 17,677 instances split into a training set (17,058 instances) and a test set (619 instances), annotated with over 77k criteria and 471k checking points in total.

\subsubsection{Rubric Construction Pipeline}
We design a highly automated pipeline to translate complex scientific generation requirements into structured, quantifiable rubrics. Given a problem statement and its associated reference image, the pipeline proceeds through four successive stages.

\noindent\textbf{Multimodal Constraint Extraction.}
We leverage Doubao-Seed-2.0-Lite \cite{seed2.0} to jointly parse the textual problem statement and reference image, extracting core scientific indicators that govern the figure's correctness. These indicators are formulated as high-level evaluation criteria.

\noindent\textbf{Atomic Point Expansion.}
Each criterion is further decomposed into a set of fine-grained checking points, where each point is formulated as an objective binary question (Yes/No). This decomposition reduces macroscopic quality judgements into atomic, ``unit-test-like'' verification signals, effectively mitigating subjective evaluation bias.

\noindent\textbf{Weighted Score Allocation.}
The total score for each rubric is strictly capped at 10.0. Weights are dynamically allocated across criteria based on the structural complexity and the risk of visual failure, ensuring that scientifically critical aspects receive proportional emphasis.

\noindent\textbf{Quality Refinement and Denoising.}
We conduct manual filtering combined with automated sanity checks to discard sub-optimal annotations, ensuring high interpretability and scientific consistency across the dataset.

\subsubsection{Data Characteristics}
Table~\ref{tab:rubric_stats} summarizes the annotation statistics across the training and test splits. Both splits exhibit consistent fine-grained annotation density: the test set averages 4.57 criteria and 29.40 checking points per figure, while the training set averages 4.35 criteria and 26.58 checking points. Across both splits, the World Knowledge subset exhibits the highest evaluation density (33.98 and 35.81 points per sample, respectively), providing rich supervisory signals for the rubric generator to internalize complex cross-disciplinary scientific knowledge.

\begin{table}[htbp]
    \centering
    \caption{Annotation statistics of \textit{SciGen-Rubric} across training and test splits. C/S: criteria per sample; P/C: points per criterion; P/S: points per sample.}
    \label{tab:rubric_stats}
    \setlength{\tabcolsep}{4.5pt}
    \renewcommand{\arraystretch}{1.15}
    \begin{tabular}{l|ccc|ccc}
        \toprule
        \multirow{2}{*}{\textbf{Category}} & \multicolumn{3}{c|}{\textbf{Test}} & \multicolumn{3}{c}{\textbf{Train}} \\
        \cmidrule{2-7}
        & \textbf{C/S} & \textbf{P/C} & \textbf{P/S} & \textbf{C/S} & \textbf{P/C} & \textbf{P/S} \\
        \midrule
        Instruction Following & 3.55 & 4.45 & 15.82 & 3.41 & 4.61 & 15.68 \\
        World Knowledge & 4.97 & 6.84 & 33.98 & 5.67 & 6.31 & 35.81 \\
        Multi. Reasoning & 4.21 & 6.64 & 27.98 & 4.14 & 6.79 & 28.11 \\
        \midrule
        \textbf{Average} & \textbf{4.57} & \textbf{6.43} & \textbf{29.40} & \textbf{4.35} & \textbf{6.12} & \textbf{26.58} \\
        \bottomrule
    \end{tabular}
\end{table}

\subsection{Training Details of Rubric Generator}
\label{app:rubric_generator}

Given a scientific image generation or editing instruction $I$ together with its associated reference image $V^{+}$, the rubric generator is trained to emit a hierarchical rubric $R = \{(c_k, P_k, w_k)\}_{k=1}^{K}$, where each element pairs a high-level criterion $c_k$ with a set of binary checking points $P_k = \{p_{k,j}\}_{j=1}^{n_k}$ and a criterion weight $w_k$.
We directly use the training split of \textit{SciGen-Rubric} described above (17,058 instances) as the SFT corpus. We instantiate the rubric generator from Qwen3-VL-8B-Instruct \cite{qwen3-vl} and perform full-parameter supervised fine-tuning by optimizing:
\begin{equation}
\mathcal{L}_{\text{SFT}}(\theta) = -\mathbb{E}_{(x,y) \sim \mathcal{D}_{\text{SFT}}} \left[ \log \pi_\theta(y \mid x) \right],
\end{equation}
where $x$ denotes the instruction with the reference image and $y$ denotes the target rubric.

\subsection{Performance of Advanced MLLMs}
\label{app:rubric_performance}

\begin{table*}[htbp]
\centering
\caption{\label{tab:rubric_performance} Performance comparison on \textit{SciGen-Rubric} across three capability domains. RR, NHR, and SF1 denote \textit{RubricRecall}, \textit{Non-HallucinationRate}, and \textit{StructuralF1}, respectively. Instruction, Reasoning, and Knowledge correspond to Instruction Following, Multidisciplinary Reasoning, and World Knowledge. The overall scores are macro-averaged across domains. The best and second-best performance in each column are highlighted in \textbf{bold} and \underline{underlined}, respectively.}
\resizebox{\textwidth}{!}{
\setlength{\tabcolsep}{3pt}
\renewcommand{\arraystretch}{1.2}
\begin{tabular}{l|ccc|ccc|ccc|ccc}
\toprule[1pt]
\multirow{2}{*}{\textbf{Model}}
& \multicolumn{3}{c|}{\textbf{Instruction}}
& \multicolumn{3}{c|}{\textbf{Reasoning}}
& \multicolumn{3}{c|}{\textbf{Knowledge}}
& \multicolumn{3}{c|}{\textbf{Overall}} \\
\cmidrule{2-13}
& \textbf{RR} & \textbf{NHR} & \textbf{SF1}
& \textbf{RR} & \textbf{NHR} & \textbf{SF1}
& \textbf{RR} & \textbf{NHR} & \textbf{SF1}
& \textbf{RR} & \textbf{NHR} & \textbf{SF1} \\
\midrule
Seed-2.0-Pro
& 65.25 & 73.74 & 68.56
& 54.91 & 68.56 & 59.72
& 57.00 & \underline{78.77} & 64.98
& 59.05 & 73.69 & 64.42 \\
Gemini-3-Flash
& 59.79 & 72.90 & 64.99
& 46.29 & 65.51 & 53.11
& 48.99 & 76.84 & 58.86
& 51.69 & 71.75 & 58.99 \\
Qwen3.6-Plus
& 57.72 & 72.43 & 63.26
& 50.24 & \underline{70.34} & 57.34
& 51.42 & 78.07 & 61.12
& 53.13 & 73.61 & 60.57 \\
Qwen3.5-397B
& 56.65 & 75.53 & 63.87
& 43.27 & 67.92 & 51.19
& 48.26 & \textbf{80.99} & 59.34
& 49.39 & \underline{74.81} & 58.13 \\
MiMo-V2-Omni
& 53.71 & 68.52 & 59.04
& 44.68 & 66.53 & 52.15
& 47.88 & 75.86 & 57.76
& 48.76 & 70.30 & 56.32 \\
Qwen3.5-9B
& 51.22 & 44.45 & 46.98
& 53.10 & 58.56 & 54.61
& 51.67 & 50.09 & 49.23
& 52.00 & 51.03 & 50.27 \\
\midrule
Ours-SFT
& \underline{77.89} & \underline{81.66} & \underline{79.29}
& \underline{64.58} & 70.09 & \underline{66.43}
& \underline{71.06} & 71.05 & \underline{70.64}
& \underline{71.18} & 74.27 & \underline{72.12} \\
Ours-SFT+RL
& \textbf{79.20} & \textbf{82.77} & \textbf{80.48}
& \textbf{67.81} & \textbf{73.84} & \textbf{69.86}
& \textbf{71.80} & 71.73 & \textbf{71.31}
& \textbf{72.94} & \textbf{76.11} & \textbf{73.88} \\
\bottomrule[1pt]
\end{tabular}
}
\end{table*}

\subsubsection{Evaluation Metrics}
We evaluate rubric generation quality by measuring how well the generated checking points cover expert-annotated gold rules while avoiding hallucinated constraints.
For each sample, let $G=\{g_j\}_{j=1}^{M}$ denote the set of gold rules and let $\hat{P}=\{\hat{p}_i\}_{i=1}^{N}$ denote the flattened set of generated checking points obtained by concatenating all $P_k$ from the predicted rubric $R$.
We define $m_{i,j}=\mathrm{match}(\hat{p}_i, g_j)\in\{0,1\}$, where $m_{i,j}=1$ indicates that $\hat{p}_i$ semantically matches $g_j$, determined by the semantic matching prompt in Appendix~\ref{app:rubric_eval_prompt} with GLM-4.7 \cite{glm4.7} (thinking enabled) as the judge.
We report three metrics:

\vspace{0.5em}
\noindent\textbf{Rubric Recall (RR)} measures the proportion of gold rules covered by any generated checking point:
\begin{equation}
\mathrm{RR}
=
\frac{1}{M}
\sum_{j=1}^{M}
\mathbb{I}\!\left[
\sum_{i=1}^{N} m_{i,j} > 0
\right].
\end{equation}

\noindent\textbf{Non-Hallucination Rate (NHR)} measures the proportion of generated checking points matched to at least one gold rule. We define the hallucination rate $\mathrm{HR}$ and take its complement:
\begin{equation}
\mathrm{HR}
=
\frac{1}{N}
\sum_{i=1}^{N}
\mathbb{I}\!\left[
\sum_{j=1}^{M} m_{i,j} = 0
\right],
\quad
\mathrm{NHR} = 1 - \mathrm{HR}.
\end{equation}

\noindent\textbf{Structural F1 (SF1)} is the harmonic mean of RR and NHR, balancing coverage and precision:
\begin{equation}
\mathrm{SF1}
=
\frac{2 \cdot \mathrm{RR} \cdot \mathrm{NHR}}
{\mathrm{RR}+\mathrm{NHR}}.
\end{equation}

All aggregate metrics are computed from the underlying matched-rule statistics before rounding; the table reports NHR rather than HR, so higher values indicate better performance for all metrics.

\subsubsection{Results and Analysis}

Table~\ref{tab:rubric_performance} reports domain-wise rubric generation quality on the test set of \textit{SciGen-Rubric}. We highlight the following findings:
\textbf{(i) The rubric generator with RL achieves the best overall performance.}
Ours-SFT+RL attains the highest scores across all three metrics, reaching 72.94 RR, 76.11 NHR, and 73.88 SF1 in the macro-averaged overall result. Compared with Ours-SFT, the GRPO-based refinement consistently improves rubric quality in every domain, with the largest gain appearing in the Reasoning domain (+3.43 SF1). This indicates that verifier feedback is particularly effective for improving rubrics that require structural and logical reasoning over scientific diagrams.
\textbf{(ii) Frontier MLLMs exhibit a clear coverage--precision trade-off.}
Across baseline models, we observe a systematic tension between coverage and conservativeness. For example, Qwen3.5-397B obtains the highest Knowledge-domain NHR (80.99), suggesting fewer hallucinated checking points, yet its substantially lower RR (48.26) indicates that it misses a large portion of expert-annotated rules. Conversely, smaller models such as Qwen3.5-9B achieve moderate RR but suffer from severe hallucination (51.03 overall NHR). In contrast, the trained rubric generator maintains both high coverage and high non-hallucination rate simultaneously, yielding the strongest SF1 in all three domains.
\textbf{(iii) These results justify using the rubric generator's outputs as process-level supervision for SciGen-Verifier}, as the generated rubrics faithfully recover the majority of expert-annotated rules while introducing minimal hallucinated constraints.

\section{More Experiment Details}

\subsection{Training Details of SciGen-Verifier}
\label{app:verifier_training}

\begin{figure}[t]
  \centering
  \includegraphics[width=\textwidth]{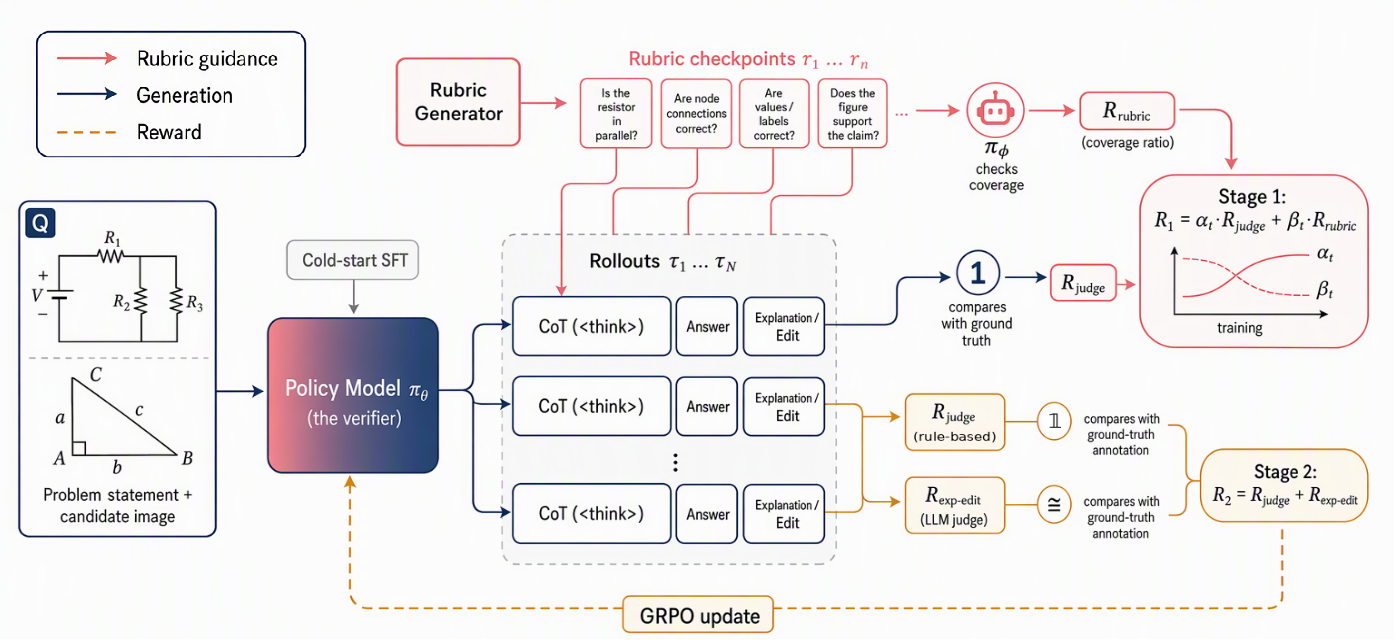}
  \caption{Overview of the SciGen-Verifier training pipeline. Starting from the cold-start SFT checkpoint, the verifier policy $\pi_\theta$ samples $N$ rollouts $\tau_1,\dots,\tau_N$ ($N{=}K{=}8$) for each problem statement and candidate image, each consisting of a \texttt{<think>} trace, a binary answer, and an explanation with an editing instruction. In Stage~1, the rubric generator produces instance-specific checkpoints $r_1,\dots,r_n$, a coverage checker $\pi_\phi$ measures how many of them are addressed in the reasoning trace ($\mathcal{R}_{\text{rubric}}$), and this process reward is blended with the rule-based judgement reward $\mathcal{R}_{\text{judge}}$ through the mastery-driven coefficients $\alpha_t$ and $\beta_t$ (Eq.~\ref{eq:total-reward}). In Stage~2, $\mathcal{R}_{\text{judge}}$ is combined with the LLM-judged explanation-editing reward $\mathcal{R}_{\text{exp-edit}}$ (Eq.~\ref{eq:stage2-total-reward}). Both stages update $\pi_\theta$ with GRPO.}
  \label{fig:method_overview}
\end{figure}

\paragraph{Overall Pipeline.}
Figure~\ref{fig:method_overview} summarizes the complete training pipeline. Given a problem statement and a candidate image, the verifier policy $\pi_\theta$, initialized from the cold-start SFT checkpoint, samples a group of rollouts that follow the tagged output format described below. The first RL stage attaches two auxiliary components to these rollouts: the rubric generator, which produces instance-specific checkpoints $r_1,\dots,r_n$ (Appendix~\ref{app:scigen-rubric}), and a coverage checker $\pi_\phi$, which marks the checkpoints explicitly addressed in the \texttt{<think>} trace and thereby yields the coverage ratio $\mathcal{R}_{\text{rubric}}$. This process-level signal is blended with the rule-based judgement reward $\mathcal{R}_{\text{judge}}$ under the mastery-driven schedule $(\alpha_t, \beta_t)$ detailed in Appendix~\ref{appendix:dynamic-weighting}. The second RL stage drops the rubric branch and instead scores the \texttt{<explanation>} and \texttt{<edit>} fields against ground-truth annotations with LLM judges, producing $\mathcal{R}_{\text{exp-edit}}$, while $\mathcal{R}_{\text{judge}}$ is retained to protect judgement accuracy. Both stages optimize $\pi_\theta$ with GRPO; only the reward composition changes between them.

\paragraph{Output Format.}
SciGen-Verifier follows the same tripartite interface used in \textit{SciGen-Verify}: given a scientific image generation or editing instruction and its candidate image, it outputs a binary judgement, a supporting explanation, and an optional editing instruction. During both SFT and RL, responses are constrained to an XML-style template consisting of \texttt{<think>}, \texttt{<answer>}, \texttt{<explanation>}, and \texttt{<edit>} fields. The \texttt{<think>} field contains the intermediate reasoning trace, while the latter three fields provide the structured outputs used for parsing and evaluation. For benchmark inference and baseline evaluation, we use the verifier prompt shown in Appendix~\ref{app:verifier_prompt}.

\paragraph{Cold-start SFT.}
We initialize SciGen-Verifier from Qwen3-VL-8B-Instruct \cite{qwen3-vl} and perform full-parameter SFT for two epochs on the curated 48,311-instance corpus described in Sec.~\ref{sec:sft}. The corpus is approximately balanced between true and false instances and contains 19,678 instruction-following, 20,367 multidisciplinary-reasoning, and 8,266 world-knowledge examples. To mitigate data imbalance across categories, we apply data-level temperature sampling during SFT, where the sampling quota of each subset is proportional to $q_i \propto n_i^{1/T}$ with $T=1.5$. The resulting SFT checkpoint serves as the initialization for all subsequent GRPO stages.

\paragraph{Train--Test Separation.}
Although the SFT corpus and \textit{SciGen-Verify} share upstream sources, they are split at the source-sample level rather than the instance level: all instances derived from the same source problem or image, including its perturbed positive and negative variants produced by the visual and textual perturbation strategies in Sec.~\ref{sec:dataset}, are assigned to exactly one of the two sets. Consequently, neither a benchmark sample nor any of its variants can flow into the training data, and the split is mutually exclusive by construction. The same rule is applied to the RL query pool and the \textit{SciGen-Rubric} training split.

\paragraph{Stage 1 RL.}
For the first RL stage, we use the SFT model to perform $K=8$ rollouts for each candidate query and retain 15\textit{K} partially successful queries whose binary judgements are correct in $k \in \{1,\dots,7\}$ rollouts. This selection removes trivial samples that are already solved in all rollouts while preserving instances that still expose meaningful reward variation. The policy is optimized with GRPO \cite{grpo} using the rubric-guided reward in Eq.~\ref{eq:total-reward}, which combines judgement correctness and rubric coverage through dynamic coefficients. This stage primarily strengthens evidence exploration and scientific constraint coverage in the generated \texttt{<think>} traces.

\paragraph{Stage 2 RL.}
The second RL stage further polishes the structured outputs using 685 fully annotated examples with answer, explanation, and editing-instruction supervision. Unlike Stage 1, which emphasizes process-level exploration, this stage directly optimizes the semantic alignment of the final explanation and editing instruction through the outcome reward in Eq.~\ref{eq:stage2-total-reward}. To avoid degrading the judgement ability learned in the first stage, answer correctness remains part of the reward throughout this stage. This two-stage design first establishes robust scientific verification and then improves the actionable feedback required by \textit{SciGen-Verify} evaluation.

\paragraph{Training Infrastructure.}
All SFT and RL experiments are implemented based on a fork of ms-swift, with additional support for data-level temperature sampling and difficulty-aware GRPO sampling. In SFT, temperature sampling with $T=1.5$ rebalances category exposure across instruction following, multidisciplinary reasoning, and world knowledge. In RL, the sampler first expands the candidate prompt pool for rollout generation and then selects prompt groups with higher intra-group reward variance for optimization. This mechanism is independent of the reward definition and improves the density of informative GRPO updates.

\subsubsection{Details of Rubric-guided Reward}
\label{appendix:dynamic-weighting}
We provide additional details on the dynamic weighting strategy used in Eq.~\ref{eq:total-reward}. The central challenge in combining judgement supervision and rubric-guided process supervision lies in their asymmetric reliability. While judgement correctness directly reflects the verifier's optimization objective, rubric coverage provides only a proxy signal for reasoning quality and may occasionally reward partially relevant or incomplete reasoning traces.
To preserve this asymmetry, we treat judgement correctness as the dominant optimization signal and use rubric coverage primarily as an auxiliary shaping reward. Specifically, the first-stage reward is formulated as:
\begin{equation}
\mathcal{R}_{1}(\tau)
=
\alpha_t \mathcal{R}_{\text{judge}}(\tau)
+
\beta_t \mathcal{R}_{\text{rubric}}(\tau),
\label{eq:appendix-total-reward}
\end{equation}
where $\alpha_t$ and $\beta_t$ dynamically control the relative contribution of judgement correctness and rubric coverage throughout training.

To adaptively balance exploration and optimization, we estimate the training mastery using a moving average of recent judgement accuracy:
\begin{equation}
m_t \in [0,1],
\label{eq:mastery}
\end{equation}
where larger values indicate stronger verifier performance. Based on the estimated mastery, the weighting coefficients are scheduled as:
\begin{equation}
\alpha_t
=
\alpha_{\max}
m_t^{\rho},
\label{eq:alpha}
\end{equation}
\begin{equation}
\beta_t
=
\beta_{\max}
(1-m_t)^{\rho},
\label{eq:beta}
\end{equation}
where $\alpha_{\max}$ and $\beta_{\max}$ denote the maximum weighting coefficients and $\rho > 0$ controls the sharpness of the curriculum schedule.
To ensure that judgement correctness remains the principal optimization target, we further impose:
\begin{equation}
\alpha_{\max}
>
\beta_{\max},
\label{eq:dominance}
\end{equation}
which prevents the auxiliary rubric-guided reward from dominating the optimization objective.

This adaptive weighting induces a curriculum-like transition during reinforcement learning. In early training, when judgement accuracy remains low (small $m_t$), rubric-guided supervision receives relatively larger weight, encouraging the verifier to explore relevant scientific checkpoints, visual evidence, and disciplinary constraints. As training progresses and verifier accuracy improves, optimization gradually shifts toward judgement correctness, allowing the policy to refine decision-making while retaining process-level guidance.
Conceptually, this formulation can be interpreted as a soft relaxation of outcome-gated process supervision. Rather than explicitly suppressing process rewards for incorrect predictions, the verifier progressively shifts from exploration-oriented reasoning supervision toward judgement-centric optimization as policy competence improves. We found this simpler adaptive formulation to be empirically more stable while preserving the intended asymmetry between process-level supervision and outcome correctness.

\subsection{Baselines Implementation}
\label{app:baselines}
We compare SciGen-Verifier against 15 leading MLLMs, including proprietary models GPT-6-Astra, Gemini-3-Flash, Gemini-3.1-Flash-Lite, Gemini-3.1-Pro, Gemini-3.5-Flash \cite{gemini3}, Seed-1.8\cite{seed1.8}, Seed-2.0-Pro \cite{seed2.0}, MiMo-V2-Omni \cite{mimo-vl}, Qwen3.8-Max, GLM-5.3-Flash, and DeepSeek-V4-Flash; open-source models Qwen3.5-397B \cite{qwen3.5}, Qwen3-VL-8B-Thinking, and Qwen3-VL-235B-Thinking \cite{qwen3-vl}, as well as OmniVerifier-7B \cite{omniverifier}, a multimodal verifier designed for visual verification.
Table~\ref{tab:baseline_models} shows their details.

To ensure fair comparison, all baseline MLLMs are evaluated under the same input-output protocol as SciGen-Verifier. Given the problem statement and candidate image, each model is instructed with the verifier prompt in Appendix~\ref{app:verifier_prompt} to make a binary judgement and provide an explanation; when the predicted answer is false, it is additionally required to generate an editing instruction. We keep the prompt template consistent across proprietary APIs, open-source MLLMs, and the OmniVerifier-7B baseline, except for minor formatting wrappers required by different inference interfaces.

\providecommand{\modeliconlink}[3]{\raisebox{-1.5pt}{\includegraphics[height=1.05em]{#1}}\ \href{#2}{#3}}

\begin{table*}[t!]
    \centering
    \small
    \setlength{\tabcolsep}{5pt}
    \resizebox{0.95\textwidth}{!}{
    \begin{tabular}{@{}l|ll@{}}
    \toprule
        \textbf{MLLMs} & \textbf{Role} & \textbf{Link} \\
    \midrule
        \multicolumn{3}{c}{\emph{Open-Source Models}} \\
    \midrule
    Qwen3.5-397B
    & Large-scale open-source baseline
    & \modeliconlink{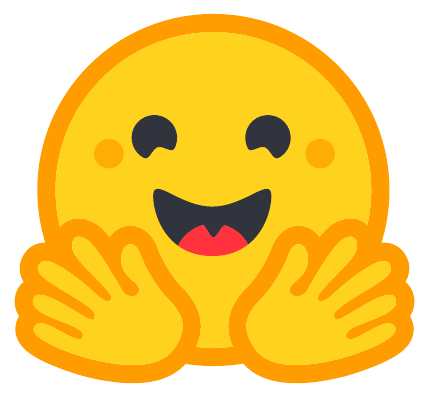}{https://huggingface.co/Qwen/Qwen3.5-397B-A17B}{Qwen/Qwen3.5-397B-A17B} \\

    Qwen3-VL-8B-Thinking
    & Same-scale thinking baseline
    & \modeliconlink{fig/icon/hf-logo.pdf}{https://huggingface.co/Qwen/Qwen3-VL-8B-Thinking}{Qwen/Qwen3-VL-8B-Thinking} \\

    Qwen3-VL-235B-Thinking
    & Large-scale thinking baseline
    & \modeliconlink{fig/icon/hf-logo.pdf}{https://huggingface.co/Qwen/Qwen3-VL-235B-A22B-Thinking}{Qwen/Qwen3-VL-235B-A22B-Thinking} \\

    OmniVerifier-7B
    & Generated-image verification baseline
    & \modeliconlink{fig/icon/hf-logo.pdf}{https://huggingface.co/comin/OmniVerifier-7B}{comin/OmniVerifier-7B} \\

    \midrule
        \multicolumn{3}{c}{\emph{Proprietary Models}} \\
    \midrule
    GPT-6-Astra
    & Strong general MLLM baseline
    & \modeliconlink{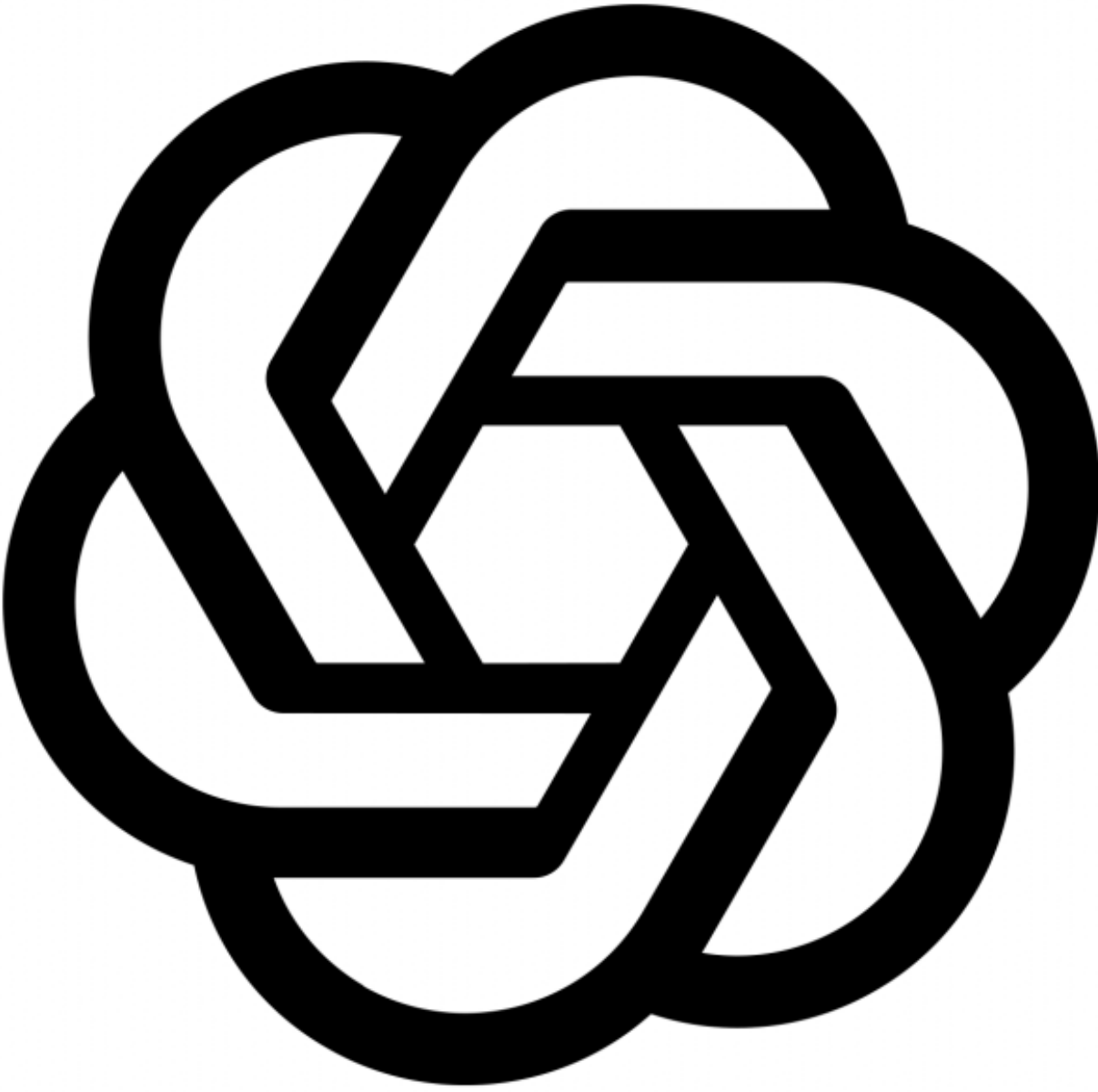}{https://openai.com/}{OpenAI model page} \\

    Gemini-3-Flash
    & General visual reasoning baseline
    & \modeliconlink{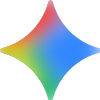}{https://ai.google.dev/gemini-api/docs/models}{Gemini API models} \\

    Gemini-3.1-Flash-Lite
    & General visual reasoning baseline
    & \modeliconlink{fig/icon/google-logo.pdf}{https://ai.google.dev/gemini-api/docs/models}{Gemini API models} \\

    Gemini-3.1-Pro
    & Strong general MLLM baseline
    & \modeliconlink{fig/icon/google-logo.pdf}{https://ai.google.dev/gemini-api/docs/models}{Gemini API models} \\

    Gemini-3.5-Flash
    & General visual reasoning baseline
    & \modeliconlink{fig/icon/google-logo.pdf}{https://ai.google.dev/gemini-api/docs/models}{Gemini API models} \\

    Seed-1.8
    & General visual reasoning baseline
    & \modeliconlink{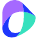}{https://seed.bytedance.com/}{SEED model page} \\

    Seed-2.0-Pro
    & Strong general MLLM baseline
    & \modeliconlink{fig/icon/seed-logo.pdf}{https://seed.bytedance.com/}{SEED model page} \\

    MiMo-V2-Omni
    & Omni-modal reasoning baseline
    & \modeliconlink{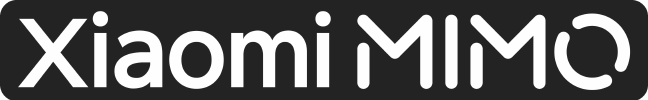}{https://mimo.mi.com/}{Xiaomi MiMo model page} \\

    Qwen3.8-Max
    & Strong general MLLM baseline
    & \modeliconlink{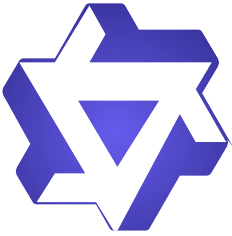}{https://www.alibabacloud.com/en/product/modelstudio}{Alibaba Cloud Model Studio} \\

    GLM-5.3-Flash
    & General visual reasoning baseline
    & \modeliconlink{fig/icon/hf-logo.pdf}{https://huggingface.co/zai-org}{Z.ai (Zhipu) HF organization} \\

    DeepSeek-V4-Flash
    & Experimental vision-language baseline
    & \modeliconlink{fig/icon/hf-logo.pdf}{https://huggingface.co/deepseek-ai/DeepSeek-V4-Flash-Vision-Exp}{deepseek-ai/DeepSeek-V4-Flash-Vision-Exp} \\
    \bottomrule
    \end{tabular}}
    \caption{The details of the baseline models in our comparison.}
    \label{tab:baseline_models}
\end{table*}

\paragraph{Decoding and Parsing.}
For each model, we parse the final response into \texttt{answer}, \texttt{explanation}, and \texttt{edit} fields. If the model does not strictly follow the XML-style template, we apply a conservative rule-based parser to extract the first unambiguous binary judgement and the corresponding free-form rationale. Responses without a valid binary judgement are treated as incorrect for $\text{Acc}_a$, and their explanation and editing-instruction scores are consequently set to zero by the cascading protocol. For positive samples, the \texttt{<edit>} field is not used in evaluation; for negative samples, a missing or non-actionable editing instruction is scored by the editing-instruction judge as invalid.

\paragraph{Evaluation.}
All models are evaluated on the same \textit{SciGen-Verify} benchmark and scored with the hierarchical metrics defined in Appendix~\ref{sec:appendix_eval}. For $\text{Acc}_e$ and $\text{Acc}_i$, we use the same model-based judges and the same ground-truth annotations for all systems. The exact judge prompts are provided in Appendix~\ref{app:judge_prompts}. This unified protocol prevents baseline-specific post-processing from altering the relative comparison.

\begin{table*}[!t]
\centering
\caption{\label{tab:tts_results} Iterative test-time scaling (TTS) results with different verifiers serving as the online critic. The value denotes the relaxed evaluation score of GenExam benchmark \cite{genexam}. We choose the Wan2.6-Image for TTS generation.}
\small
\setlength{\tabcolsep}{4pt}
\resizebox{0.95\textwidth}{!}{
\begin{tabular}{lcccccccccc}
\toprule[1pt]
\textbf{Verifier} & \textbf{Step 1} & \textbf{Step 2} & \textbf{Step 3} & \textbf{Step 4} & \textbf{Step 5} & \textbf{Step 6} & \textbf{Step 7} & \textbf{Step 8} & \textbf{Step 9} & \textbf{Step 10} \\
\midrule[1pt]
Gemini-3.5-Flash & 59.4 & 58.9 & 61.9 & 62.6 & 63.5 & 63.2 & 64.1 & 64.5 & 64.4 & 64.9 \\
Qwen3.8-Max & 59.0 & 58.3 & 61.5 & 62.0 & 62.9 & 63.4 & 63.2 & 63.9 & 64.1 & 64.4 \\
Seed-2.0-Pro & 59.2 & 55.6 & 60.8 & 61.5 & 59.4 & 62.6 & 62.3 & 61.9 & 63.0 & 62.6 \\
\midrule
SciGen-Verifier-8B & 58.7 & 57.7 & 61.3 & 61.0 & 62.4 & 63.0 & 63.0 & 63.1 & 63.2 & 63.5 \\
\bottomrule[1pt]
\end{tabular}}
\end{table*}

\section{More Experimental Analysis}
\subsection{Exploration of Reinforcement Learning}
\label{app:rl_exploration}

We further analyze how reinforcement learning contributes to SciGen-Verifier and rubric generator. For SciGen-Verifier, the central issue is how to expose the model to queries that are sufficiently challenging and how to prevent auxiliary rubric coverage from overwhelming judgement correctness. For rubric generator, the key question is whether verifier feedback can provide a meaningful signal for improving the discriminability of generated checking points.

\subsubsection{RL Data Selection for SciGen-Verifier.}
Following Sec.~\ref{sec:rl}, we sample each candidate query $K=8$ times with the SFT verifier and compute the number of correct binary judgements. Consistently correct queries offer limited learning signal due to minimal reward diversity, while consistently incorrect queries often induce noisy supervision with weak preference contrast. We therefore retain partially successful queries, namely those with $k \in \{1,\dots,7\}$ correct rollouts, and construct a 15\textit{K}-query RL pool. This selection makes each GRPO group more likely to contain both successful and failed responses, thereby improving reward contrast during policy optimization.

\subsubsection{Stage-wise RL for SciGen-Verifier.}
As shown in Table~\ref{tab:rl_strategies}, SFT alone establishes basic verification ability but remains insufficient for generating faithful corrections. Stage 1 RL improves the reasoning and judgement metrics by rewarding explicit coverage of rubric checkpoints in the \texttt{<think>} trace. However, because this stage mainly supervises evidence exploration, its improvement on editing instruction accuracy remains limited. Stage 2 RL directly adds outcome-level supervision for explanation and editing instruction, leading to the strongest overall performance. These results support the curriculum design: reliable scientific verification should first be strengthened through process-level exploration and then aligned with textual feedback quality.

\begin{figure}[t]
    \centering
    \includegraphics[width=\linewidth]{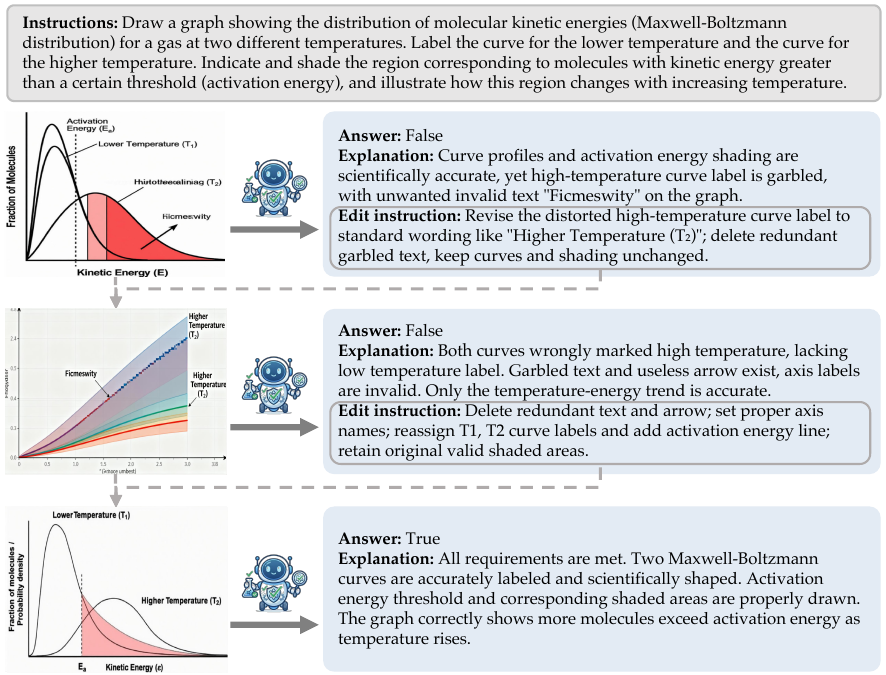}
    \caption{Sequential TTS case demonstrations with SciGen-Verifier feedback.}
    \label{tts_case}
\end{figure}

\subsubsection{RL with SciGen-Verifier Feedback.}
During GRPO, we employ SciGen-Verifier as the reward source for training rubric generator. The Verifier's principal capability is to judge whether an image satisfies a given instruction, and each binary checking point emitted by rubric generator is itself an image-verification instruction. Point-wise evaluation therefore directly reuses the judgement capability already acquired by the Verifier. Feeding the entire rubric $R = \{(c_k, P_k, w_k)\}_{k=1}^{K}$ in a single query would push the input off the Verifier's training distribution; we therefore adopt a point-wise verification protocol that pairs each checking point $p_{k,j} \in P_k$ individually with the candidate image and queries the Verifier for its satisfaction status.

For each training instance, we construct one positive image--instruction pair that satisfies the underlying scientific constraints and one contrastive negative pair constructed from the corresponding training data. The negative pair uses the same visual asset with a modified instruction, making the image valid for the positive pair but invalid for the negative pair. rubric generator first emits $R$, after which the Verifier evaluates every point $p_{k,j}$ against each pair under the point-wise protocol.

Let $z_{k,j}(V)\in\{0,1\}$ denote the Verifier's binary result for checking point $p_{k,j}$ on image $V$, where $z_{k,j}(V)=1$ indicates that the point is satisfied. The criterion-level score is computed as
\begin{equation}
\phi_k(V,R)
=
\frac{1}{n_k}
\sum_{j=1}^{n_k}
z_{k,j}(V),
\end{equation}
where $n_k=|P_k|$. The image-level score is then aggregated by criterion weights:
\begin{equation}
S(V,R)
=
\sum_{k=1}^{K}
w_k\,\phi_k(V,R),
\end{equation}
where $\sum_{k=1}^{K} w_k=10.0$. This produces a positive score $S^{+}$ and a negative score $\bar{S}^{-}$ for the paired contrastive instance. The reward decomposes into a discriminability term and a format term,
\begin{equation}
\mathcal{R} = \lambda_{s}\, \mathcal{R}^{\mathrm{score}} + \mathcal{R}^{\mathrm{fmt}},
\label{eq:rubricator-reward}
\end{equation}
with the score reward defined piecewise as
\begin{equation}
\mathcal{R}^{\mathrm{score}} =
\begin{cases}
-1.0, & S^{+} \le 7.0, \\[4pt]
\dfrac{S^{+} - \bar{S}^{-}}{10.0}, & S^{+} > 7.0.
\end{cases}
\label{eq:rubricator-score}
\end{equation}
The positive-score threshold guards the basic faithfulness of $R$: if the positive pair cannot exceed $7.0$, the rubric has failed to capture the constraints that the valid image necessarily satisfies and is treated as structurally invalid, receiving a fixed negative reward. Above the threshold, the margin $S^{+} - \bar{S}^{-}$ governs the reward and directly quantifies discriminability---a wider margin indicates that the checking points sharply localize the distinguishing features between the positive and negative pairs, whereas a vanishing margin reflects a rubric that no longer separates the two and is penalized accordingly.
The format term $\mathcal{R}^{\mathrm{fmt}}$ enforces structural validity of the emitted rubric: the response must be valid JSON encoding $R = \{(c_k, P_k, w_k)\}_{k=1}^{K}$, where each element carries a weight $w_k$ subject to $\sum_{k=1}^{K} w_k = 10$. Full compliance yields $1.0$; any violation yields $0.0$.

\subsection{Application of Test-time Scaling}
\label{app:tts}
Beyond offline verification, SciGen-Verifier can be used as an online critic for iterative test-time scaling (TTS) of scientific image generation. Given an original generation prompt, we first obtain an initial image from Wan2.6-t2i \cite{tongyi2026wan26} and query SciGen-Verifier with the verifier prompt in Appendix~\ref{app:verifier_prompt}. If the verifier predicts \textit{true}, the procedure stops early and returns the current image. If the verifier predicts \textit{false}, the system decides between regeneration and local editing based on the predicted editing instruction: editing instructions containing three or more correction items or global-regeneration keywords trigger regeneration, while localized corrections are sent to Wan2.6-image \cite{tongyi2026wan26} for image editing. The edited or regenerated image is then passed back to the verifier, and the process is repeated for at most 10 steps. We do not introduce an additional verifier prompt for TTS; the same verifier prompt is reused to obtain the answer, explanation, and editing instruction.

As shown in Table~\ref{tab:tts_results} and Fig.~\ref{tts_case}, verifier-guided refinement improves the relaxed score from 58.7 at the initial generation step to 63.5 after up to ten steps. Although the score does not monotonically increase at every intermediate step, the overall trend demonstrates that structured verification feedback can progressively correct scientific image generation errors. For reference, we also plug frontier MLLMs into the same loop as the critic. Gemini-3.5-Flash and Qwen3.8-Max reach 64.9 and 64.4 after ten steps, only 1.4 and 0.9 points above our 8B verifier, while Seed-2.0-Pro exhibits a more volatile trajectory and ends at 62.6. This indicates that SciGen-Verifier delivers online-critic utility on par with much larger proprietary models at a fraction of the inference cost. Note that this relaxed score is used only for the TTS application and is not directly comparable to the hierarchical $\text{Acc}_a$, $\text{Acc}_e$, and $\text{Acc}_i$ metrics used in \textit{SciGen-Verify}.

\clearpage
\onecolumn
\section{Prompt Templates}
\label{app:prompt_templates}

\newenvironment{scipromptbox}[1]{%
\begin{tcolorbox}[title = {#1}, breakable, colback=gray!3, colframe=black!65, colbacktitle=black!70, coltitle=white, fonttitle=\bfseries\small, fontupper=\small, boxrule=0.6pt, arc=2pt, left=1.5mm, right=1.5mm, top=1mm, bottom=1mm]
\raggedright
\setlength{\parindent}{0pt}
\setlength{\parskip}{0.35em}
}{\end{tcolorbox}}

\newcommand{\promptfield}[1]{\vspace{0.3em}\noindent\textbf{#1}\par}
\newcommand{\prompttag}[1]{\texttt{\detokenize{#1}}}
\newcommand{\promptvar}[1]{\texttt{\{#1\}}}

\subsection{Annotation Prompts}
\label{app:annotation_prompts}

The following prompt templates summarize the AI-assisted annotation procedure for positive and negative examples in \textit{SciGen-Verify}. Instance-specific variables such as \promptvar{instruction} and \promptvar{edit\_explanation} are replaced for each example.

\begin{scipromptbox}{Positive Explanation Annotation Prompt}
\promptfield{Role}
You are a Senior Expert-Level Scientific Image Analyst. Your mission is to provide a technical justification for the consistency between a ``Generated Image'' and its ``Input Requirements.''

\promptfield{Objective}
The provided image and instruction are already consistent. Your task is to perform a detailed visual-textual mapping to explain \textbf{why} they align, focusing on specific scientific evidence and visual markers within the image.

\promptfield{Inputs}
\textbf{Input Requirements (Ground Truth):} \promptvar{instruction}

\textbf{Visual Input:} The last image provided in the conversation.

\promptfield{Analysis Guidelines}
\begin{enumerate}
    \item \textbf{Evidence-Based Mapping:} Identify specific visual markers (e.g., data trends, chemical bonds, axis scales, or diagram components) that directly fulfill the requirements in the \promptvar{instruction}.
    \item \textbf{Scientific Interpretation:} Translate visual elements into technical confirmations (e.g., ``The downward slope of the curve confirms the `negative correlation' required'').
    \item \textbf{No Redundancy:} Do not simply repeat the instruction. Describe what is \textbf{visually present} that proves the instruction was followed.
\end{enumerate}

\promptfield{Mandatory Output Format}
\textbf{Consistency Analysis.} Provide a concise, professional explanation that bridges the text requirements and the visual evidence.
\begin{itemize}
    \item \textbf{Format:} Must be enclosed within \prompttag{<explanation> </explanation>} tags.
    \item \textbf{Requirement:} Use bullet points for multiple markers to ensure scannability.
\end{itemize}

\promptfield{Example}
\prompttag{<explanation>}\par
Consistency is confirmed through three key visual markers:
\begin{enumerate}
    \item The molecular structure correctly displays the tetrahedral geometry at the chiral center as specified in the requirements.
    \item The color-coding of the heat map strictly follows the `viridis' scale mentioned in the instruction, with the legend range correctly normalized from 0 to 1.
    \item The X-axis correctly spans from 0 to 100 ms with the 10 ms interval markers explicitly rendered and aligned with the data points.
\end{enumerate}
\prompttag{</explanation>}
\end{scipromptbox}
\clearpage

\begin{scipromptbox}{Modified-Instruction Negative Annotation Prompt}
\promptfield{Role}
You are a Senior Expert-Level Scientific Image Auditor. Your task is to verify an existing image against a newly updated instruction, using the provided modification rationale as your primary audit guide.

\promptfield{Context}
\begin{itemize}
    \item \textbf{Target Instruction:} The new goal the image must satisfy.
    \item \textbf{Modification Rationale:} A detailed explanation of what has changed and why the instruction was updated.
    \item \textbf{Existing Image:} The last image provided in the conversation, which was generated based on a previous, now outdated version of the instruction.
\end{itemize}

\promptfield{Audit Task}
\begin{enumerate}
    \item \textbf{Analyze the Rationale:} Identify the specific delta (changes) described in the ``Modification Rationale.''
    \item \textbf{Verify Inconsistency:} Inspect the ``Existing Image'' to find where it still follows the old logic and thus contradicts the ``Target Instruction.''
    \item \textbf{Bridge the Gap:} Determine what visual edits are necessary to align the image with the new requirements.
\end{enumerate}

\promptfield{Inputs}
\textbf{Target Instruction:} \promptvar{instruction}

\textbf{Modification Rationale:} \promptvar{edit\_explanation}

\promptfield{Mandatory Output Format}
\textbf{1. Conflict Analysis.} Identify precisely why the current image is now incorrect. Directly list visual elements that fail to meet the ``Target Instruction'' based on the ``Modification Rationale.''
\begin{itemize}
    \item \textbf{Format:} Must be enclosed within \prompttag{<explanation> </explanation>} tags.
    \item \textbf{Style:} Concise, evidence-based, no filler words.
\end{itemize}

\textbf{2. Actionable Editing Instruction.} Provide a step-by-step technical guide to transform the existing image into the target state.
\begin{itemize}
    \item \textbf{Format:} Must be enclosed within \prompttag{<edit> </edit>} tags.
    \item \textbf{Requirement:} Instructions must be concrete (e.g., ``Change the bar color from blue to green'', ``Replace the legend label `A' with `B''').
\end{itemize}

\promptfield{Example}
\prompttag{<explanation>}\par
The ``Modification Rationale'' specifies a shift from a 2D plot to a 3D perspective to show depth. The existing image remains a 2D scatter plot, missing the Z-axis (Depth) and the corresponding data points required by the Target Instruction.\par
\prompttag{</explanation>}\par
\prompttag{<edit>}\par
\begin{enumerate}
    \item Project the 2D scatter plot into a 3D coordinate system.
    \item Add a Z-axis labeled ``Depth (nm)''.
    \item Update the data points to include the third dimension values.
\end{enumerate}
\prompttag{</edit>}
\end{scipromptbox}
\clearpage

\begin{scipromptbox}{Code-Based Negative Annotation Prompt with Reference}
You are a senior expert-level scientific image analysis model. There is a misalignment between the requirements and the visual output. Compare the ``Image to be Fixed'' with the ``Correct Reference Image.''

\promptfield{Visual Input}
\begin{itemize}
    \item \textbf{Image to be Fixed:} The \textbf{second to last} image in the conversation.
    \item \textbf{Correct Reference Image:} The \textbf{last} image in the conversation.
\end{itemize}

\promptfield{Input Requirements}
\promptvar{instruction}

\promptfield{Task \& Output Format (Mandatory)}
\begin{enumerate}
    \item \textbf{Conflict Analysis:} Be extremely concise. Directly list the missing elements or visual discrepancies found in the ``Image to be Fixed'' when compared to the reference. This part \textbf{MUST} be enclosed within \prompttag{<explanation> </explanation>} tags.
    \item \textbf{Actionable Editing Instruction:} Provide targeted steps to correct the image. This part \textbf{MUST} be enclosed within \prompttag{<edit> </edit>} tags.
\end{enumerate}

\promptfield{Example}
\prompttag{<explanation>}\par
The image is missing the error bars shown in the reference, the data trend is linear instead of polynomial, and the title font size is too small.\par
\prompttag{</explanation>}\par
\prompttag{<edit>}\par
\begin{enumerate}
    \item Add vertical error bars to all data points.
    \item Apply a 2nd-order polynomial fit.
    \item Increase title font size to 14pt.
\end{enumerate}
\prompttag{</edit>}
\end{scipromptbox}
\clearpage

\begin{scipromptbox}{Code-Based Negative Annotation Prompt without Reference}
You are a senior expert-level scientific image analysis model. The generated image fails to satisfy the specified requirements.

\promptfield{Visual Input}
The ``Existing Image'' is the \textbf{last image} in the conversation.

\promptfield{Input Requirements}
\promptvar{instruction}

\promptfield{Task \& Output Format (Mandatory)}
\begin{enumerate}
    \item \textbf{Defect Analysis:} Be extremely concise and critical. Directly list the elements from the instruction that are missing or incorrect in the image. Use a format like ``None of the...: there is no X, no Y, and Z is wrong.'' This part \textbf{MUST} be enclosed within \prompttag{<explanation> </explanation>} tags.
    \item \textbf{Actionable Editing Instruction:} Provide clear instructions to repair the image. This part \textbf{MUST} be enclosed within \prompttag{<edit> </edit>} tags.
\end{enumerate}

\promptfield{Example}
\prompttag{<explanation>}\par
None of the required annotations are present: there is no scale bar at the bottom right, no arrows pointing to the cell nucleus, and the stain color is blue instead of the required purple.\par
\prompttag{</explanation>}\par
\prompttag{<edit>}\par
\begin{enumerate}
    \item Add a 10$\mu$m scale bar to the bottom right.
    \item Draw arrows pointing to the high-intensity regions.
    \item Change the colormap to purple.
\end{enumerate}
\prompttag{</edit>}
\end{scipromptbox}
\clearpage

\subsection{Verifier Inference Prompt}
\label{app:verifier_prompt}

The following prompt is used for SciGen-Verifier inference and for baseline evaluation. The placeholder \promptvar{instruction} is replaced by the task-specific scientific image generation or editing instruction, and the candidate image is attached at the \prompttag{<image>} position.

\begin{scipromptbox}{SciGen-Verifier Inference Prompt}
You are an expert scientific image verification model. Your task is to evaluate whether a generated image correctly follows the given instruction. You must analyze the correctness of the generated image step by step, then provide a final judgement.

\promptfield{Task Types}
\begin{itemize}
    \item \textbf{Instruction Following:} Drawing/editing scientific diagrams based on explicit requirements (e.g., ``draw a line at x=5'', ``add axis labels'', ``mark point P'').
    \item \textbf{Multidisciplinary Reasoning:} Creating figures that require logical deduction (e.g., plotting calculated data points, geometric constructions based on given conditions).
    \item \textbf{World Knowledge:} Illustrating scientific concepts from domain knowledge (e.g., ``draw Earth's internal layers'', ``sketch a plant cell structure'').
\end{itemize}

\promptfield{Instruction}
\promptvar{instruction}

\promptfield{Generated Image}
\prompttag{<image>}

\promptfield{Evaluation Rules}
\begin{enumerate}
    \item The instruction is the \textbf{ONLY} evaluation standard --- do not add requirements not mentioned in the instruction.
    \item All equivalent implementations are correct (e.g., equivalent parallel circuits, either solid or dashed lines are acceptable if no style is specified).
    \item Ignore all stylistic aspects unless explicitly required (e.g., colors, line thickness, font styles, marker shapes).
    \item All explicitly requested elements must be present with correct values, positions, and labels.
    \item The logical/scientific conclusion must be correct.
\end{enumerate}

\promptfield{Judgement}
\begin{itemize}
    \item \textbf{true:} All requirements fulfilled (equivalent implementations allowed), scientifically/logically correct.
    \item \textbf{false:} Any required element missing, any value/position wrong, or any scientific/logical error.
\end{itemize}

\promptfield{Output}
\prompttag{<answer>true/false</answer>}\par
\prompttag{<explanation>Explain why the judgement is correct or incorrect.</explanation>}\par
\prompttag{<edit>If <answer> is false, provide concrete editing instructions to fix the image. If <answer> is true, do not output this tag.</edit>}
\end{scipromptbox}
\clearpage

\subsection{Judge Prompts}
\label{app:judge_prompts}

The following judge prompts are used to compute $\text{Acc}_e$ and $\text{Acc}_i$. The explanation judge compares the predicted explanation with the ground-truth explanation, while the edit judge compares the predicted editing instruction with the ground-truth editing instruction.

\begin{scipromptbox}{Explanation Evaluation Prompt}
You are a strict evaluator for a scientific image verification system. Your task is to compare the ``Predicted Explanation'' generated by a model against the ``Ground Truth Explanation''.

\promptfield{Context (Instruction)}
\promptvar{instruction}

\promptfield{Ground Truth Judgement}
\promptvar{gt\_judgement}

\promptfield{Ground Truth Explanation (Standard Answer)}
\promptvar{gt\_explanation}

\promptfield{Predicted Explanation (To be Evaluated)}
\promptvar{pred\_explanation}

\promptfield{Task}
Determine whether the ``Predicted Explanation'' captures the same verification rationale as the ``Ground Truth Explanation''. For positive samples, it should identify the same key visual evidence supporting consistency with the instruction. For negative samples, it should identify the same core discrepancy, violated constraint, or scientific error.

\promptfield{Judgement Criteria}
\begin{itemize}
    \item Return \texttt{true} if the predicted explanation captures the same key supporting evidence for a positive sample, or the same core error for a negative sample.
    \item Return \texttt{false} if the prediction is irrelevant, contradicts the instruction or ground truth, misses the main rationale, or introduces unsupported requirements.
\end{itemize}

\promptfield{Output Format (JSON only)}
\{\par
\quad \texttt{"result"}: \texttt{<boolean, true or false>}\par
\}
\end{scipromptbox}
\clearpage

\begin{scipromptbox}{Editing Instruction Evaluation Prompt}
You are a senior scientific image editor. You are reviewing a suggested ``Editing Instruction'' to fix an incorrect image based on a ``Generation Context'' and the provided images.

\promptfield{Background Context}
\begin{itemize}
    \item The \textbf{Generation Context} below includes the original instruction and the generated image result (represented by \prompttag{<image>}).
    \item The generated image is \textbf{incorrect} and contains errors or deviations from the instruction requirements.
    \item A \textbf{Ground Truth Editing Instruction} is provided as a reference, which describes how to correctly fix the errors in the generated image.
\end{itemize}

\promptfield{Generation Context (Instruction \& Incorrect Image)}
\promptvar{instruction}

\textbf{Incorrect Image:} \prompttag{<image>}

\promptfield{Ground Truth Editing Instruction (Standard Reference for Fixing)}
\promptvar{gt\_edit}

\promptfield{Predicted Editing Instruction (To be Evaluated)}
\promptvar{pred\_edit}

\promptfield{Task}
Judge whether the ``Predicted Editing Instruction'' is a valid and correct fix for the incorrect image. Compare it with the ``Ground Truth Editing Instruction'' and the visual context provided in the generation context.
\begin{itemize}
    \item Return \texttt{true} if the predicted editing instruction is semantically consistent with the ground truth or effectively fixes the issue.
    \item Return \texttt{false} if the predicted editing instruction is irrelevant, incorrect, contradictory to the ground truth, or fails to address the core problem.
\end{itemize}

\promptfield{Output Format (JSON only)}
\{\par
\quad \texttt{"result"}: \texttt{<boolean, true or false>}\par
\}
\end{scipromptbox}
\clearpage

\subsection{Rubric Generation Prompt}
\label{app:rubric_generation_prompt}

The following prompt is used to generate structured rubrics for \textit{SciGen-Rubric}. The prompt requires each rubric to be a valid JSON object with criteria, atomic yes/no checking points, and criterion scores summing to 10.0.

\begin{scipromptbox}{Rubric Generation Prompt}
You are a professional rubric generation expert. Your task is to generate precise, detailed scoring rubrics for evaluating AI-generated scientific diagrams.

You will be given the original image generation instruction.

Your output must be a valid JSON object strictly following this format:

\{\par
\quad \texttt{"rubric"}: [\par
\quad\quad \{\par
\quad\quad\quad \texttt{"criterion"}: \texttt{"<The general evaluation criterion - a high-level aspect to assess>"},\par
\quad\quad\quad \texttt{"points"}: [\texttt{"<A specific yes/no sub-question>"}, \ldots],\par
\quad\quad\quad \texttt{"score"}: \texttt{<criterion\_score>}\par
\quad\quad \},\par
\quad\quad \ldots\par
\quad ]\par
\}

\promptfield{Rubric Structure Rules}
\begin{itemize}
    \item Total score across all criteria \textbf{MUST} sum to exactly 10.0 points.
    \item Allocate scores based on the number of distinct sub-requirements, points of failure, and structural importance. Criteria with more sub-elements or higher risk of visual error should receive proportionally higher scores. Criteria that are foundational or frequently violated should also receive higher scores.
    \item All score values must be rounded to 2 decimal places.
\end{itemize}

\promptfield{Criterion Design Guidelines}
\begin{itemize}
    \item Each criterion should cover \textbf{ONE} distinct, non-overlapping evaluation dimension.
    \item Each point should be a single, atomic, answerable yes/no question about that dimension.
    \item Points should cover: (1) presence/correctness of required elements, (2) absence of errors or unwanted extra elements, (3) structural/logical relationships.
    \item Questions must be answerable by visually inspecting the generated image against the original instruction.
    \item Use clear, specific language. Avoid vague terms like ``properly'' or ``adequately'' without definition.
\end{itemize}

\promptfield{Important}
\begin{itemize}
    \item Only output the JSON object. Do \textbf{NOT} add any explanation, preamble, or additional text outside the JSON.
    \item The JSON must be parseable directly without any markdown formatting.
\end{itemize}
\end{scipromptbox}
\clearpage

\subsection{Rubric Evaluation Prompt}
\label{app:rubric_eval_prompt}

The following prompt is used to semantically align candidate rubric rules with gold standard rules when computing \textit{RubricRecall}, \textit{Non-HallucinationRate}, and \textit{StructuralF1}. The prompt enforces strict scope matching so that overly broad or vague candidate rules are not counted as valid hits.

\begin{scipromptbox}{Rubric Semantic Matching Prompt}
You are an expert evaluator for the Rubric benchmark. Your task is to perform a batch semantic alignment between a set of ``Candidate Rubric Rules'' and a set of ``Gold Standard Rules''.

\promptfield{Instruction}
Compare each Candidate Rule against the entire Gold Rules list. Determine if there is a semantic match.

\promptfield{Strict Matching Criteria}
A ``Hit'' (YES) requires:
\begin{enumerate}
    \item \textbf{Specific Intent Match:} The Candidate Rule must check the \textbf{EXACT SAME} constraint as the Gold Rule (e.g., if Gold checks ``Structure'', Candidate must check ``Structure'', not just ``Quality'').
    \item \textbf{Scope Match:} The Candidate Rule must not be significantly broader or vaguer than the Gold Rule.
\end{enumerate}

\promptfield{Automatic Rejection Criteria (NO)}
\begin{itemize}
    \item \textbf{Vague vs Specific:} If Candidate says ``Is the explanation good/detailed?'' and Gold says ``Does it mention Concept X?'', this is NO.
    \item \textbf{Different Dimension:} If Candidate checks ``Content'' and Gold checks ``Structure/Formatting'', this is NO.
    \item \textbf{Partial Overlap:} If Candidate checks ``Relevance'' but maps it to a Gold Rule about ``Completeness'', this is NO (unless the correct Gold Rule is missing).
\end{itemize}

\promptfield{Evaluation Policy (Must Follow)}
Semantic equivalence means the Candidate Rule would accept and reject the same set of responses as the Gold Rule in practice. Any broadening, weakening, or generalization of constraints counts as a scope mismatch. Do \textbf{NOT} combine partial overlaps across multiple Gold Rules to justify a YES. If ``hit'' is NO, return an empty list for \texttt{hit\_gold\_rule\_indices}.

\promptfield{Critical JSON Rules}
\begin{enumerate}
    \item Wrap \textbf{ALL} keys and string values in double quotes. Never use single quotes.
    \item Escape all literal double quotes inside strings using backslash, e.g., \texttt{\textbackslash"Word\textbackslash"}.
    \item Escape all newlines inside strings with \texttt{\textbackslash n}. Do not output literal line breaks inside strings.
    \item Separate items in lists and dictionaries with commas. Do not miss the comma.
\end{enumerate}

\promptfield{Input Data}
\textbf{Gold Rules List:} \promptvar{gold\_rules}\par
\textbf{Candidate Rules to Evaluate:} \promptvar{candidate\_rules}

\promptfield{Output Format}
Output a strictly valid JSON object where the key \texttt{"match\_results"} contains a list of objects, one for each Candidate Rule in the exact order provided. Do not include markdown formatting or any other commentary.

\promptfield{Example}
\{\par
\quad \texttt{"match\_results"}: [\par
\quad\quad \{ \texttt{"candidate"}: \texttt{"Rule 1 text"}, \texttt{"hit"}: \texttt{"YES"}, \texttt{"hit\_gold\_rule\_indices"}: [0, 2] \},\par
\quad\quad \{ \texttt{"candidate"}: \texttt{"Rule 2 text"}, \texttt{"hit"}: \texttt{"NO"}, \texttt{"hit\_gold\_rule\_indices"}: [] \}\par
\quad ]\par
\}
\end{scipromptbox}
\clearpage

\onecolumn
\section{More Cases}

\subsection{Cases of SciGen-Verifier}

\begin{tcolorbox}[title = {Illustrative Examples of SciGen-Verifier in World Knowledge Domain}, breakable, fontupper=\small, fonttitle=\small]

\medskip
\begin{center}
\includegraphics[height=3.2cm]{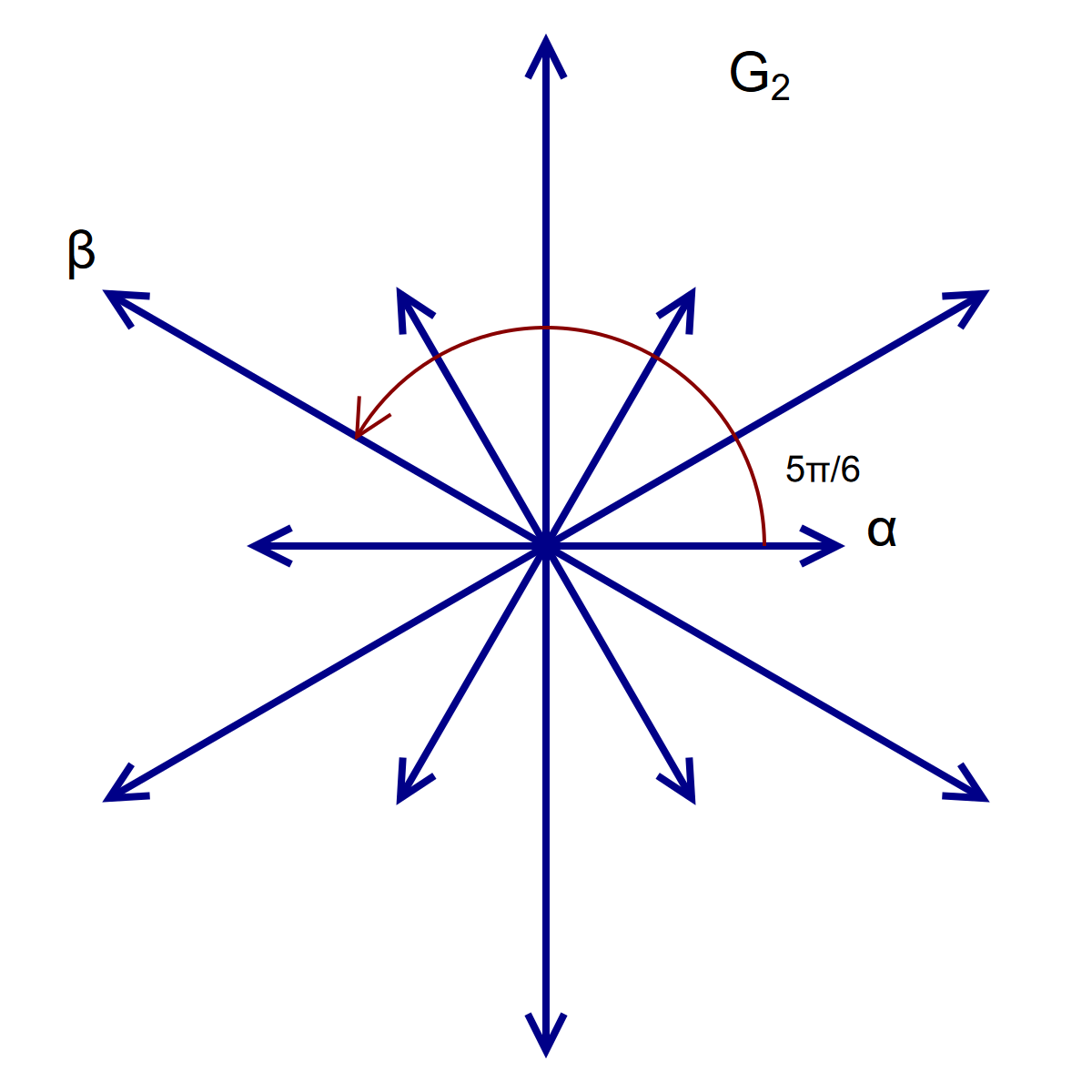}
\end{center}

\medskip
\textbf{Problem: } Create a diagram of the discipline based on the following description: \\This is a geometric structure diagram showing the Root system of type $G_2$. All vectors (roots) in the diagram originate from the origin of the coordinate system and are arranged radially on a two-dimensional plane. The root system consists of 12 roots, composed of 6 short roots and 6 long roots, with a length ratio of $1:\sqrt{3}$. Short roots and long roots are alternately distributed, and the angle between adjacent roots is $\pi/6$ (30°).\\\\The key elements in the diagram are as follows:\\1. \textbf{Vector distribution}:\\   - 6 \textbf{short roots} point in the directions of 0°, 60°, 120°, 180°, 240°, and 300°, respectively.\\   - 6 \textbf{long roots} point in the directions of 30°, 90°, 150°, 210°, 270°, and 330°, respectively.\\   - All roots are represented by dark blue arrows.\\2. \textbf{Mathematical annotations}:\\   - The short root in the 0° direction is labeled with the Greek letter $\alpha$.\\   - The long root in the 150° ($5\pi/6$ radians) direction is labeled with the Greek letter $\beta$.\\   - These two roots $\alpha$ and $\beta$ form a set of \textbf{Simple roots} of the root system.\\   - There is a red curved arrow in the diagram rotating from $\alpha$ to $\beta$, annotated with the angle value $5\pi/6$, indicating the angle between these two sets of simple roots.\\3. \textbf{Text labels}:\\   - The symbol $G_2$ is noted in the upper right corner of the image, representing the type of the Lie algebra.\\4. \textbf{Overall layout}:\\   - The background is pure white, and the composition possesses a high degree of sixth-order symmetry ($D_6$ dihedral group symmetry).

\medskip
\hrule
\medskip

\raggedright
\colorbox{green!10}{\textbf{\fontsize{11pt}{10pt}\selectfont SciGen-Verifier}}

\vspace{3pt}

{\scriptsize\ttfamily
\textbf{\textcolor{blue}{<think>}}

Got it, let's go through this step by step. First, let's check all the requirements one by one.\par\medskip\par First, the root directions: short roots are supposed to be 0°, 60°, 120°, 180°, 240°, 300°. Let's look at the image: 0° (right) is short, 60° up-right is short, 120° up-left is short, 180° left is short, 240° down-left is short, 300° down-right is short. That's 6 short roots, right? Then long roots are 30°, 90°, 150°, 210°, 270°, 330°. Let's check: 30° is between 0 and 60, that's the long one, 90° up is long, 150° between 120 and 180, that's the one labeled $\beta$, right? Wait, the label $\beta$ is on 150° which is correct, $\alpha$ is 0° which is the short root labeled $\alpha$, that's correct. Then the angle between $\alpha$ and $\beta$ is 5$\pi/6$ which is what's labeled, the red curved arrow is from $\alpha$ to $\beta$, that's right. G2 is in upper right, yes, that's there.\par\medskip\par Wait, what about the length ratio? Wait, the short roots are the shorter ones? Wait no, wait wait, let's count the roots: how many total roots? 12, right 6+6=12. Let's count the arrows here: 1. right ($\alpha$, short), 2. upper right (60°, short), 3. up (90°, long), 4. upper left (120°, short), 5. left (180°, short), 6. lower left (240°, short), 7. down left (210°, long), 8. down (270°, long), 9. lower right (300°, short), 10. lower right (330°, long). Wait wait that's 10? Wait no, count again: let's list all:\\1. (0°, right) → $\alpha$: 1\\2. (30°, up-right): 2\\3. (60°, upper-right): 3\\4. (90°, up): 4\\5. (120°, upper-left): 5\\6. (150°, upper-left): 6 → $\beta$, correct.\\7. (180°, left):7\\8. (210°, lower-left):8\\9. (240°, lower-left):9\\10. (270°, down):10\\11. (300°, lower-right):11\\12. (330°, lower-right):12. Oh right 12 total. Now, are short roots the 6 directions: 0, 60,120,180,240,300. That's 6, which are exactly the shorter arrows here, yes. The 6 long are 30,90,150,210,270,330, those are the longer arrows, which matches the length ratio, right? The ratio is ~1:1.732, looks like they are longer than the short ones here.\par\medskip\par Then the angle between adjacent roots: $\pi/6$ (30°), that's correct because 12 roots equally spaced 30° apart, that's what the image has. Alternating short and long, which is correct: 0(short),30(long),60(short),90(long),120(short),150(long),180(short)... yes that's alternating, correct.\par\medskip\par Then the annotations: $\alpha$ on 0° short, $\beta$ on 150° long, that's exactly what the image has. The red curved arrow from $\alpha$ to $\beta$ labeled 5$\pi/6$, that's correct. G2 in upper right, yes. Background is white, the symmetry is sixth order dihedral, which this diagram has, all elements are present, correct positions, all required elements are there.\par\medskip\par Wait wait, did I miss anything? Let's check again. The instruction says all vectors originate from origin, arranged radially on 2D plane, yes, all start at the origin. All are dark blue arrows, yes. Short roots in those directions, long in those directions, correct. The angle between $\alpha$ and $\beta$ is 5$\pi/6$, that's correct: 150° minus 0° is 150° which is 5$\pi/6$, yes. So all requirements are fulfilled. So the judgment is true.

\textbf{\textcolor{blue}{</think>}}
}

\vspace{2pt}

\textbf{Answer:} True

\textbf{Explanation:} All the requirements specified in the instruction are correctly satisfied:\\1.  There are 12 total roots (6 short roots and 6 long roots) originating from the origin, arranged radially on a 2D plane, with alternating short/long root distribution, adjacent roots separated by 30° ($\pi/6$), which matches the required structure. The length ratio of long to short roots is correctly followed (long roots are longer than short roots, matching the $1:\sqrt{3}$ requirement).\\2.  Short roots are correctly placed at directions 0°, 60°, 120°, 180°, 240°, 300°, and long roots at 30°, 90°, 150°, 210°, 270°, 330°, all represented as dark blue arrows.\\3.  Annotations are correct: the 0° short root is labeled $\alpha$, the 150° long root is labeled  $\beta$, the angle between $\alpha$ and  $\beta$ is marked as $5\pi/6$ with a red curved arrow from $\alpha$ to  $\beta$ as required.\\4.  The $G_2$ symbol is correctly placed in the upper right corner, the background is pure white, and the diagram has the required sixth-order dihedral symmetry. All scientific and geometric requirements are met.

\vspace{6pt}

\colorbox{red!10}{\textbf{\fontsize{11pt}{10pt}\selectfont Seed-2.0-Pro}}

\vspace{2pt}

\textbf{Answer:} False

\textbf{Explanation:} The generated image does not distinguish between short and long roots: all 12 vectors are of equal length, which violates the requirement that there are 6 short roots and 6 long roots with a length ratio of $1:\sqrt{3}$ that are alternately distributed.

\textbf{Editing Instruction: } Adjust the lengths of the vectors: set the length of the 6 roots pointing to 0°, 60°, 120°, 180°, 240°, 300° (including the short root $\alpha$) to $1/\sqrt{3}$ of the length of the 6 roots pointing to 30°, 90°, 150°, 210°, 270°, 330° (including the long root  $\beta$), to meet the $1:\sqrt{3}$ length ratio requirement for short and long roots.

\end{tcolorbox}
\clearpage

\begin{tcolorbox}[title = {Illustrative Examples of SciGen-Verifier in Multidisciplinary Reasoning Domain}, breakable, fontupper=\small, fonttitle=\small]

\medskip
\begin{center}
\includegraphics[height=3.2cm]{fig/case_study/bench_case_2_compound4.png}
\end{center}

\medskip
\textbf{Problem: } Create a multiple-choice organic chemistry exam figure asking: "Which para-substituted benzoic acid is most acidic?" Draw four separate structural options of benzene rings, each bearing a carboxyl group (–COOH) at the top position and a para substituent at the bottom position. The four substituents, one per structure, must be: Br, OCH3, CN, and NO2. Arrange the four options side-by-side with clear separation, without indicating the correct answer.

\medskip
\hrule
\medskip

\raggedright
\colorbox{green!10}{\textbf{\fontsize{11pt}{10pt}\selectfont SciGen-Verifier}}

\vspace{3pt}

{\scriptsize\ttfamily
\textbf{\textcolor{blue}{<think>}}

Got it, let's go through this step by step. First, check the requirements. The instruction says: four para-substituted benzoic acid, each has COOH at top, para substituent at bottom, the four substituents are Br, OCH3, CN, NO2, arranged side by side, no indicating correct answer, it's a multiple choice question asking which is most acidic.\par\medskip\par First, check each structure: all have COOH at the top, para substituent at bottom, right? Let's list them: first (A) Br at para, second OCH3, third CN, fourth NO2. All are para positions, correct substituents, all four are there. They are arranged side by side, separate. Wait, the instruction says "without indicating the correct answer"—wait the image labels them (A), (B), (C), (D) but it doesn't say which is correct, just that don't indicate correct answer, but even with labels, does that matter? Wait no, the instruction says "without indicating the correct answer"—the labels are just identifiers, the question is still there? Wait no, wait the substituents are all correct positions. Let's confirm: COOH at top, para (bottom position) for each: yes. The four substituents are exactly Br, OCH3, CN, NO2, one each. All four structures separate side by side. The question is "which para-substituted benzoic acid is most acidic"—that's the correct prompt. The instruction says don't indicate the correct answer, the labels here are just option labels, not indicating which is correct. Wait, the instruction says "without indicating the correct answer"—so just labeling options (A) to (D) doesn't indicate which is correct, that's okay. All requirements are met: all four structures are correct, substituents are right, positions are right, arranged side by side. Scientifically, this setup is correct, the most acidic would be NO2, but the image doesn't indicate that, which follows the instruction. Wait wait, the instruction says "without indicating the correct answer"—the image doesn't mark any as correct, just has the options. So all requirements are fulfilled. So the judgment is true. Let me double check: 1. Multiple choice question asking the right question: yes, the question is implied (it's the common question, the image sets up the four options correctly). 2. Four separate structures, each COOH at top, para substituent at bottom: yes. 3. Four substituents: Br, OCH3, CN, NO2: all present, correct. 4. Arranged side by side, separated: yes. 5. No indication of correct answer: the only labels are option labels (A/B/C/D), which doesn't indicate which is correct, that's allowed. All requirements are satisfied. So answer is true.

\textbf{\textcolor{blue}{</think>}}
}

\vspace{2pt}

\textbf{Answer:} True

\textbf{Explanation:} All requirements of the instruction are fully satisfied:\\1. The question setup is correct, asking for the most acidic para-substituted benzoic acid as a multiple-choice question.\\2. Four separate benzene structures are arranged side-by-side with clear separation.\\3. Each structure has the carboxyl (-COOH) group at the top position, and one correct para substituent (Br, OCH3, CN, NO2, each in their own structure, matching the required list) at the para/bottom position.\\4. No indication of the correct answer is provided, which follows the instruction.\\All structural and content requirements are met correctly.

\vspace{6pt}

\colorbox{red!10}{\textbf{\fontsize{11pt}{10pt}\selectfont Gemini-3.1-Pro}}

\vspace{2pt}

\textbf{Answer:} False

\textbf{Explanation:} The image correctly draws the four para-substituted benzoic acid structures with the specified substituents (Br, OCH3, CN, NO2) at the bottom and the carboxyl group at the top, arranged side-by-side. However, it is missing the explicitly requested question text: `Which para-substituted benzoic acid is most acidic?'.

\textbf{Editing Instruction: } Add the text `Which para-substituted benzoic acid is most acidic?' above the chemical structures.

\end{tcolorbox}
\clearpage

\begin{tcolorbox}[title = {Illustrative Examples of SciGen-Verifier in Instruction Following Domain}, breakable, fontupper=\small, fonttitle=\small]

\medskip
\begin{center}
\begin{tikzpicture}[inner sep=0]
  \node (img) {\includegraphics[height=3.2cm]{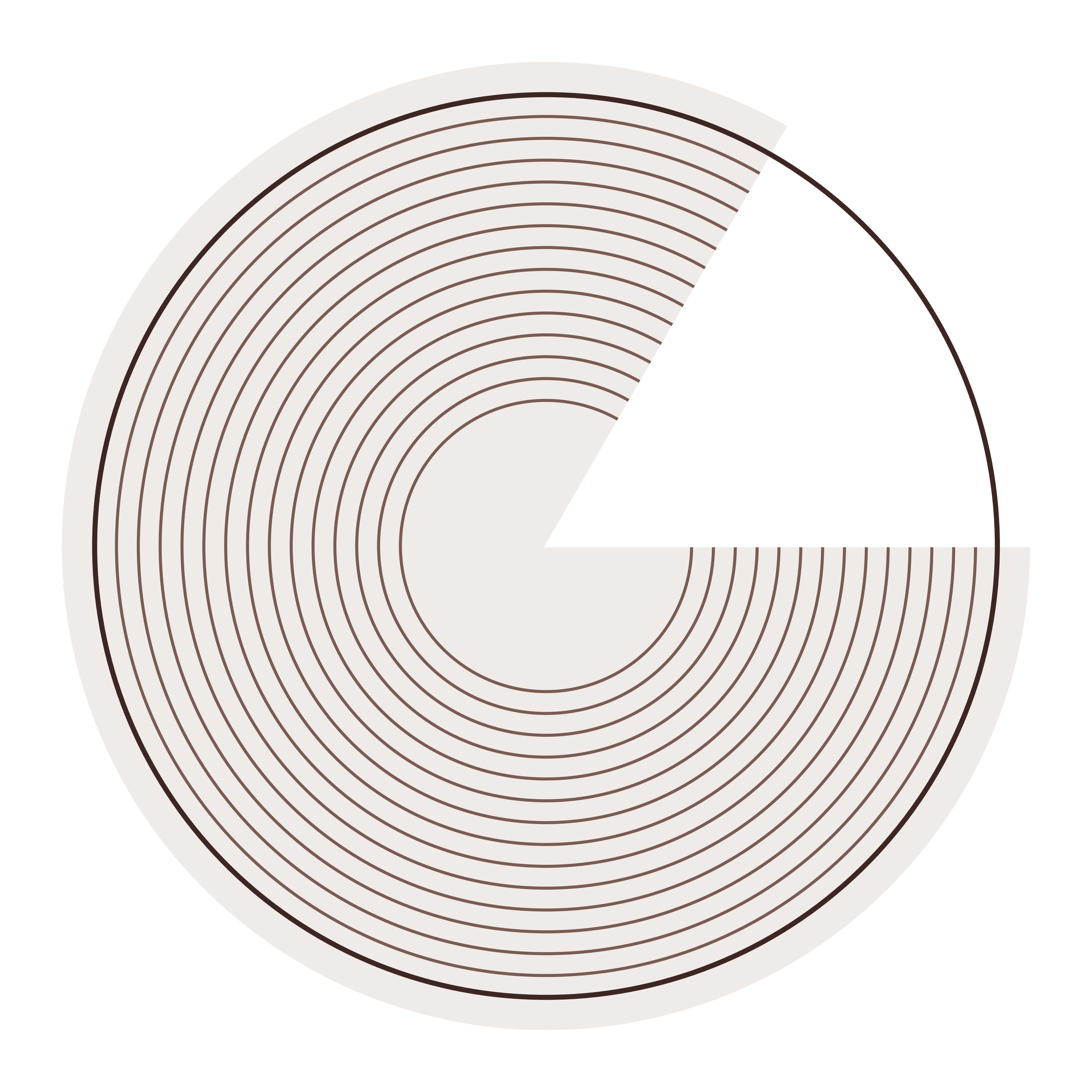}};
  \node[anchor=north,yshift=-2pt,font=\scriptsize\bfseries] at (img.north) {\textless{}Context Image\textgreater};
\end{tikzpicture}
\includegraphics[height=3.2cm]{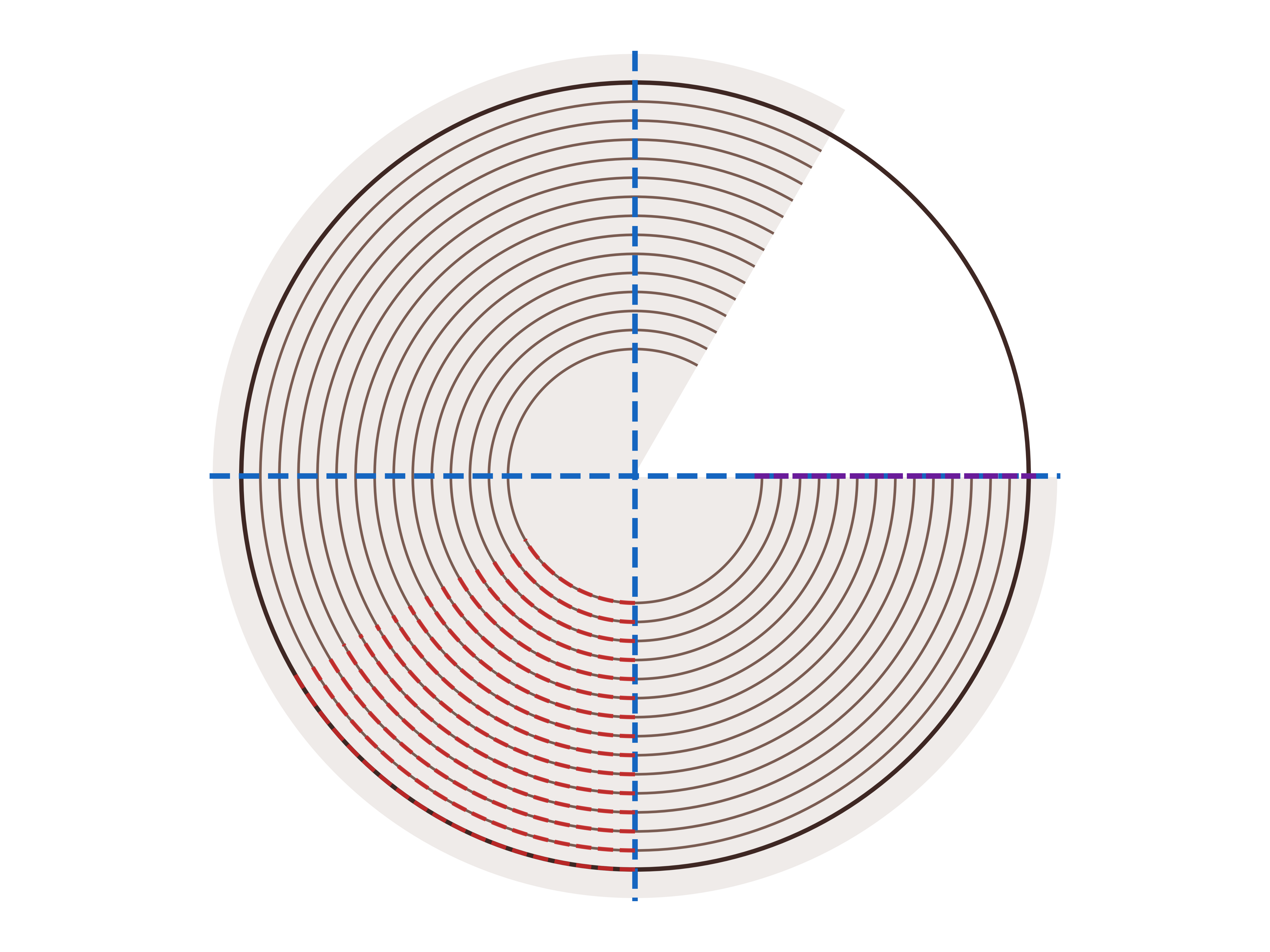}
\end{center}

\medskip
\textbf{Problem: } Edit the provided image according to the following instructions:\\1. Add two perpendicular dashed diameters through the center of the tree cross-section to establish symmetry axes\\2. Reconstruct the occluded growth rings in the hidden wedge by drawing dashed arcs to complete ring continuity\\3. Add tick marks along one unobstructed radius to mark each ring intersection for clear counting

\medskip
\hrule
\medskip

\raggedright
\colorbox{green!10}{\textbf{\fontsize{11pt}{10pt}\selectfont SciGen-Verifier}}

\vspace{3pt}

{\scriptsize\ttfamily
\textbf{\textcolor{blue}{<think>}}

Got it, let's go through each requirement step by step.\par\medskip\par First requirement: Add two perpendicular dashed diameters through the center. Looking at the generated image, the blue dashed lines are the two perpendicular diameters, crossing at the center, that's done correctly.\par\medskip\par Second requirement: Reconstruct the occluded growth rings in the hidden wedge as dashed arcs to keep continuity. Wait, wait, the occluded wedge is the white empty wedge at the top right, right? Oh, in the generated image, they added the red dashed rings in the bottom left quadrant, not the hidden (occluded, the white wedge) one. Wait no, wait original image: the wedge is the empty white part, the hidden wedge is that top right white space, right? The instruction says reconstruct the occluded growth rings in the hidden wedge (the white wedge) as dashed arcs. But the generated image added the dashed rings in the bottom left, which is already present, not the hidden wedge. Wait wait no, let me check again: original image: the wedge is the empty white part, the hidden wedge is that top right white space, right? The instruction says reconstruct the occluded growth rings in that hidden wedge (the white wedge) as dashed arcs. But the generated image put the red dashed rings in the bottom left, which is already present, not the hidden wedge. Wait wait no, wait the third requirement: add tick marks along one unobstructed radius to mark each ring intersection. Oh, that's done! The purple ticks are along the horizontal unobstructed radius, each ring intersection is marked, that's correct. Wait but the second requirement: the hidden wedge is the white empty wedge (the occluded part, the one that's missing rings). The instruction says reconstruct the occluded growth rings in that hidden wedge, draw dashed arcs there. But the generated image didn't do that, instead it added extra dashed rings in the bottom left, which is wrong. Wait wait, let me re-read instruction 2: "Reconstruct the occluded growth rings in the hidden wedge by drawing dashed arcs to complete ring continuity". The original image: the hidden wedge is the white triangular wedge (the part that's not showing the rings, right? So we need to draw the dashed arcs for each ring in that white wedge, so each ring goes from the inner center out, through that wedge, as a dashed arc. But in the generated image, they added the dashed rings in the lower left quadrant, which is not the occluded wedge. Wait wait, maybe I got that wrong. Let me check again: original image: the original image's missing part is the white wedge, the top right. So the rings are cut off there. The generated image: the rings are all there except the top right wedge, which should have the dashed rings drawn. But the generated image instead added more rings in the bottom left, which is already complete. Oh! That's a mistake. Wait but wait, the two perpendicular diameters: that's correct, they are dashed, perpendicular, through center. The tick marks: they are along the unobstructed radius, each intersection is marked, that's correct. But the second requirement is not done correctly: the occluded (hidden) wedge is the white wedge, we need to draw the dashed arcs for the rings there, but instead, the generated image added the dashed rings in the wrong place (lower left quadrant, which was already complete). Wait, is that right? Let me confirm:\par\medskip\par Original image: <img> original: the outer circle, the rings are all present except the top right wedge, that wedge has no rings. So instruction 2 says: reconstruct those occluded rings (in that hidden wedge) as dashed arcs. The generated image: the hidden wedge (top right) still has no rings. Instead, they added the dashed rings in the bottom left quadrant, which is wrong. Oh! So that's an error. Wait but wait, let me check the generated image again: the generated image has the bottom left quadrant has the red dashed rings, that's not the occluded wedge. The occluded wedge is still empty. So that's wrong. Wait but wait, maybe I misread: "occluded growth rings in the hidden wedge" — occluded meaning the wedge is the hidden part, so the rings are occluded (hidden) there, so we need to draw them. So the generated image didn't do that, it put them in the wrong place. So the judgment is false? Wait wait no, wait let's recheck:\par\medskip\par Wait original image: the center is the inner part, the white wedge is the occluded (hidden) region, right? So the rings go through that wedge, but the part in the wedge isn't drawn, so we need to draw them as dashed. The generated image: that region is still empty, no rings. The added dashed rings are in the lower-left quadrant, which is wrong. So that's a mistake. Then the third requirement: tick marks are added along the unobstructed radius, that's correct, but the second requirement is wrong. So the answer is false. Wait but wait, maybe I got the hidden wedge wrong. Is the hidden wedge the bottom left? No, original image: the outer edge has a gap at the top right, so the outer rings are missing the part in that gap, so that's the occluded wedge. Yeah. So let's structure this:\par\medskip\par Judgment is false. Explanation: 1. The two perpendicular dashed diameters through the center are correctly added, and the tick marks along the unobstructed radius are correctly added. However, the second requirement was not correctly fulfilled: the occluded/hidden wedge (the empty white wedge at the top-right of the original image, where the growth rings are missing) was not reconstructed with dashed arcs for the rings. Instead, the added dashed rings were incorrectly placed in the lower-left quadrant (which was already complete) instead of the hidden occluded wedge, which does not complete the ring continuity as requested.\par\medskip\par Then the edit instruction: Remove the incorrectly placed red dashed rings from the lower-left quadrant, and draw the missing growth ring arcs as dashed lines through the empty (occluded) top-right wedge, so each ring's arc passes through this hidden wedge as a dashed arc to complete the continuity of all growth rings.

\textbf{\textcolor{blue}{</think>}}
}

\vspace{2pt}

\textbf{Answer:} False

\textbf{Explanation:} All requirements are not correctly fulfilled:\\1. The two perpendicular dashed diameters through the circle center are correctly added, and the tick marks along the unobstructed radius are correctly placed to mark all ring intersections, which meets these two requirements.\\2. The second requirement is not satisfied: the occluded/hidden wedge is the empty white wedge at the top-right of the original image (where the growth rings are missing), but the generated image incorrectly added dashed rings to the already complete lower-left quadrant, and did not reconstruct the missing growth rings in the actual hidden occluded wedge to complete ring continuity, which is a positional error.

\textbf{Editing Instruction: } Remove the incorrectly placed red dashed rings from the lower-left quadrant, then draw the missing growth ring arcs as dashed lines through the empty occluded top-right wedge, so each ring's arc passes through this hidden wedge as a dashed arc to complete the continuity of all growth rings.

\vspace{6pt}

\colorbox{red!10}{\textbf{\fontsize{11pt}{10pt}\selectfont Gemini-3-Flash}}

\vspace{2pt}

\textbf{Answer:} True

\textbf{Explanation:} The generated image fails to follow the instructions in several ways: 1. It did not reconstruct the growth rings in the hidden wedge (the white triangular area); instead, it added red dashed arcs in the bottom-left quadrant where rings were already visible. 2. It did not add clear tick marks at each ring intersection along an unobstructed radius; the purple marks on the right horizontal radius are poorly placed and do not clearly mark the intersections for counting.

\textbf{Editing Instruction: } Reconstruct the growth rings within the white wedge using dashed arcs to complete the circles. Add distinct tick marks (small perpendicular lines) at every point where a growth ring intersects one of the unobstructed radii (e.g., the left horizontal radius or the bottom vertical radius).

\end{tcolorbox}
\clearpage

\begin{tcolorbox}[title = {Illustrative Examples of SciGen-Verifier in Multidisciplinary Reasoning Domain}, breakable, fontupper=\small, fonttitle=\small]

\medskip
\begin{center}
\begin{tikzpicture}[inner sep=0]
  \node (img) {\includegraphics[height=3.2cm]{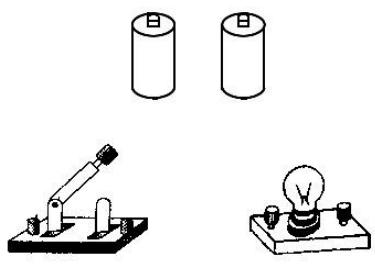}};
  \node[font=\scriptsize\bfseries,xshift=3pt] at (img.center) {\textless{}Context Image\textgreater};
\end{tikzpicture}
\includegraphics[height=3.2cm]{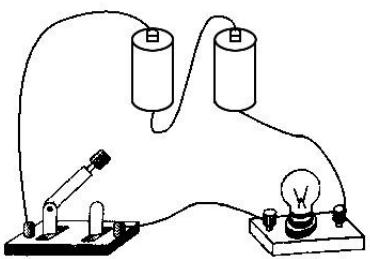}
\end{center}

\medskip
\textbf{Problem: } Please connect the components shown in the figure with pen strokes to replace wires and form a complete circuit.

\medskip
\hrule
\medskip

\raggedright
\colorbox{green!10}{\textbf{\fontsize{11pt}{10pt}\selectfont SciGen-Verifier}}

\vspace{3pt}

{\scriptsize\ttfamily
\textbf{\textcolor{blue}{<think>}}

Got it, let's work through this step by step. First, the instruction says to connect all components to form a complete circuit. Let's look at the generated image: the connections are done, it's a series circuit with two batteries, switch, and bulb. Wait, let's check the flow: starting from the negative of the first battery, to switch, then to bulb, then to second battery positive, then back to first battery negative? Wait no, wait the connection: the first battery (left) top connected to switch, then switch to bulb left terminal, bulb right terminal connected to the second battery, then the second battery's other terminal goes to the left battery's negative? Wait no, actually wait, the circuit is complete here. All components are connected, it's a closed circuit that works. Wait, is there any error? Let's see: the requirement is to connect components to form a complete circuit. The generated image has all components connected correctly, the circuit is closed, all four components (two batteries, switch, bulb) are included, connected properly with pen strokes as wires. So it's correct, right? Wait wait, let me check again. Oh yes, the connections form a complete closed circuit: the path is: left battery (+) connected to switch, switch to bulb, bulb to right battery (-), right battery (+) connected back to left battery (-)? Wait no, actually, regardless, it's a complete circuit, all components are connected, the instruction is just to form a complete circuit by connecting components. All requirements are met. So judgment is true.

\textbf{\textcolor{blue}{</think>}}
}

\vspace{2pt}

\textbf{Answer:} True

\textbf{Explanation:} The generated image correctly connects all provided components (two batteries, switch, and light bulb) with drawn wire strokes to form a complete closed circuit that satisfies the requirement of connecting components to form a complete circuit, which is logically and scientifically correct. All requirements from the instruction are fulfilled.

\vspace{6pt}

\colorbox{red!10}{\textbf{\fontsize{11pt}{10pt}\selectfont Seed-2.0-Pro}}

\vspace{2pt}

\textbf{Answer:} False

\textbf{Explanation:} The generated circuit has scientific errors: first, the negative terminal of the right battery is not connected to any component, so the right battery is not properly integrated into the circuit; second, the right terminal of the bulb is incorrectly connected to the positive terminal of the right battery, leading to conflicting polarities of the two batteries in the circuit, which makes the circuit unable to work normally.

\textbf{Editing Instruction: } Remove the wire connecting the right terminal of the bulb to the positive (top) terminal of the right battery, and connect the right terminal of the bulb to the negative (bottom) terminal of the right battery instead, so that all components form a correct complete series circuit.

\end{tcolorbox}
\clearpage

\subsection{Cases of Rubric Generator}

\begin{tcolorbox}[title = {Illustrative Examples of rubric generator in Instruction Following Domain}, breakable, fontupper=\small, fonttitle=\small]

\medskip
\begin{center}
\begin{tikzpicture}[inner sep=0]
  \node (img) {\includegraphics[height=3.2cm]{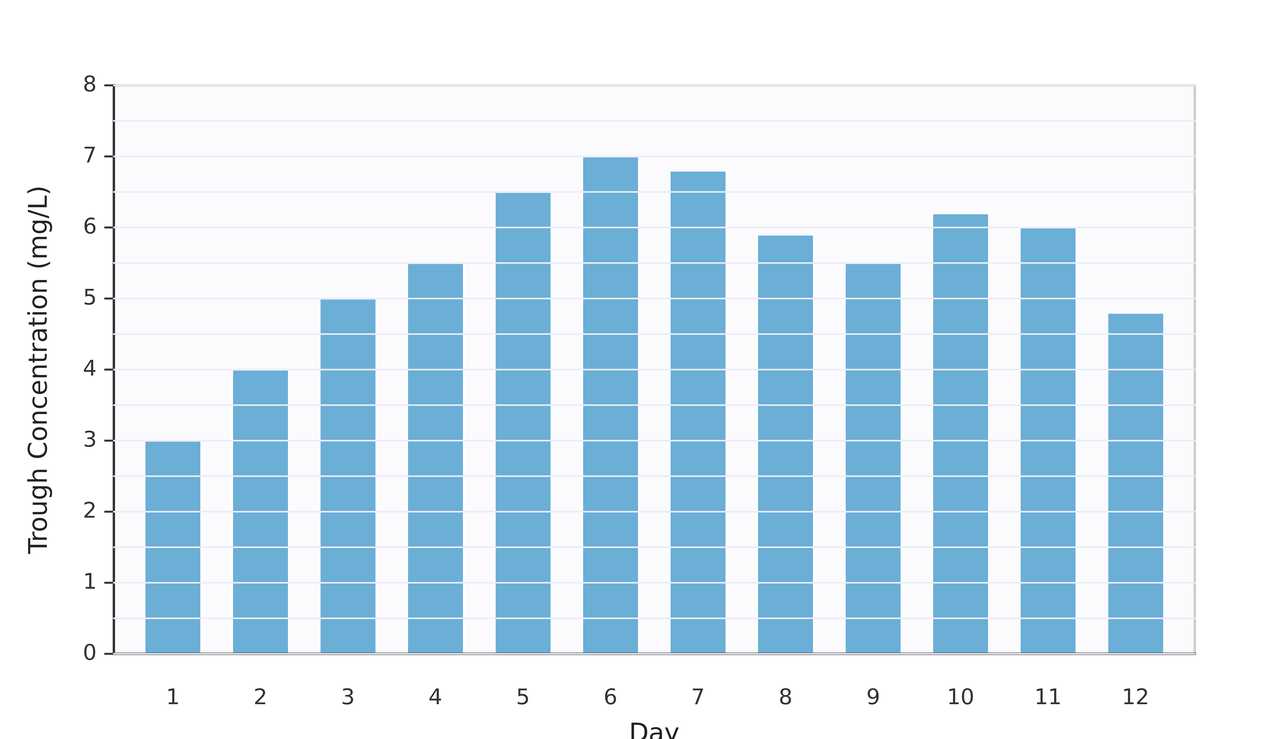}};
  \node[anchor=north,yshift=-2pt,font=\scriptsize\bfseries] at (img.north) {\textless{}Context Image\textgreater};
\end{tikzpicture}
\includegraphics[height=3.2cm]{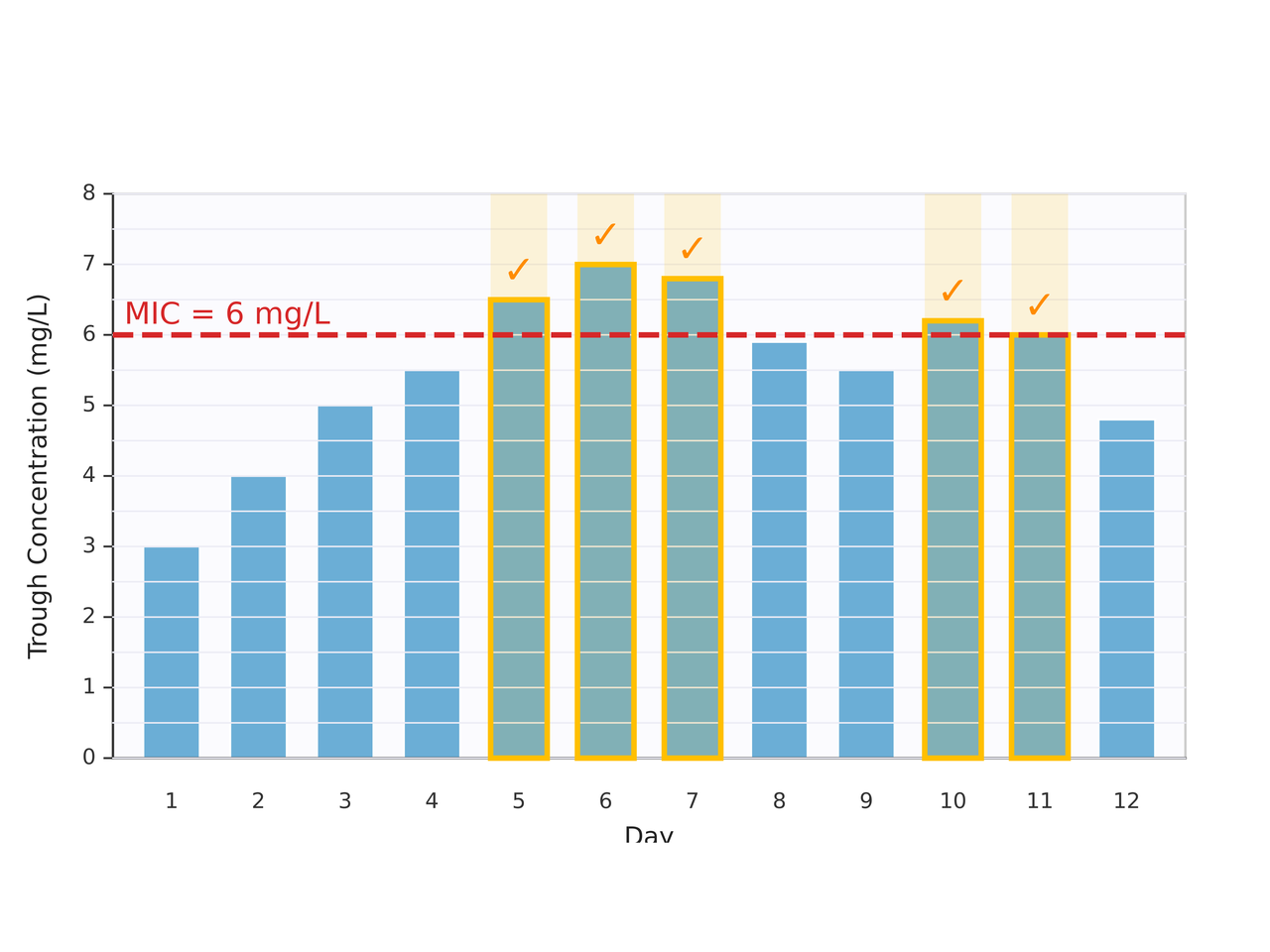}
\end{center}

\medskip
\textbf{Problem: } Edit the provided image according to the following instructions:\\1. Draw a horizontal dashed reference line at the 6 mg/L level on the y-axis, labeled ``MIC = 6 mg/L''\\2. Highlight all bars that reach or exceed the MIC line with a yellow border and background highlight\\3. Mark each highlighted bar with a check mark (\checkmark) to indicate compliance with the therapeutic threshold

\medskip
\hrule
\medskip

\noindent\textbf{Criterion 1:} \textit{A horizontal dashed reference line at the 6 mg/L level on the y-axis, labeled ``MIC = 6 mg/L'', is correctly added.}

\begin{itemize}[leftmargin=*,nosep]
    \item \textbf{Point 1.1:} Is there a horizontal reference line positioned at the 6 mg/L mark on the y-axis?
    \item \textbf{Point 1.2:} Is the horizontal reference line drawn as a dashed line?
    \item \textbf{Point 1.3:} Is the reference line correctly labeled with the text ``MIC = 6 mg/L''?
    \item \textbf{Point 1.4:} Is the label text ``MIC = 6 mg/L'' correctly spelled and formatted?
    \item \textbf{Point 1.5:} Does the image avoid placing the MIC reference line at any y-level other than 6 mg/L?
\end{itemize}

\medskip
\noindent\textbf{Criterion 2:} \textit{All bars that reach or exceed the 6 mg/L MIC line are highlighted with a yellow border and background highlight.}

\begin{itemize}[leftmargin=*,nosep]
    \item \textbf{Point 2.1:} Is every bar that reaches or exceeds the 6 mg/L MIC line properly highlighted?
    \item \textbf{Point 2.2:} Does each highlighted bar have a yellow border around its edge?
    \item \textbf{Point 2.3:} Does each highlighted bar have a yellow background highlight applied?
    \item \textbf{Point 2.4:} Are any bars that do not reach or exceed the 6 mg/L MIC line incorrectly highlighted?
    \item \textbf{Point 2.5:} Does the image avoid adding yellow highlighting to bars that do not meet the MIC threshold?
\end{itemize}

\medskip
\noindent\textbf{Criterion 3:} \textit{Each highlighted (compliant) bar that meets or exceeds the MIC line is marked with a check mark (\checkmark) to indicate compliance.}

\begin{itemize}[leftmargin=*,nosep]
    \item \textbf{Point 3.1:} Does every highlighted compliant bar have a check mark placed on it?
    \item \textbf{Point 3.2:} Is there a check mark present for every bar that meets or exceeds the MIC threshold?
    \item \textbf{Point 3.3:} Are any check marks incorrectly placed on non-highlighted non-compliant bars?
    \item \textbf{Point 3.4:} Are all the required marks correctly drawn as check (\checkmark) symbols?
    \item \textbf{Point 3.5:} Does the image avoid missing check marks on any highlighted compliant bar?
\end{itemize}

\end{tcolorbox}
\clearpage

\begin{tcolorbox}[title = {Illustrative Examples of rubric generator in World Knowledge Domain}, breakable, fontupper=\small, fonttitle=\small]

\medskip
\begin{center}
\includegraphics[height=3.2cm]{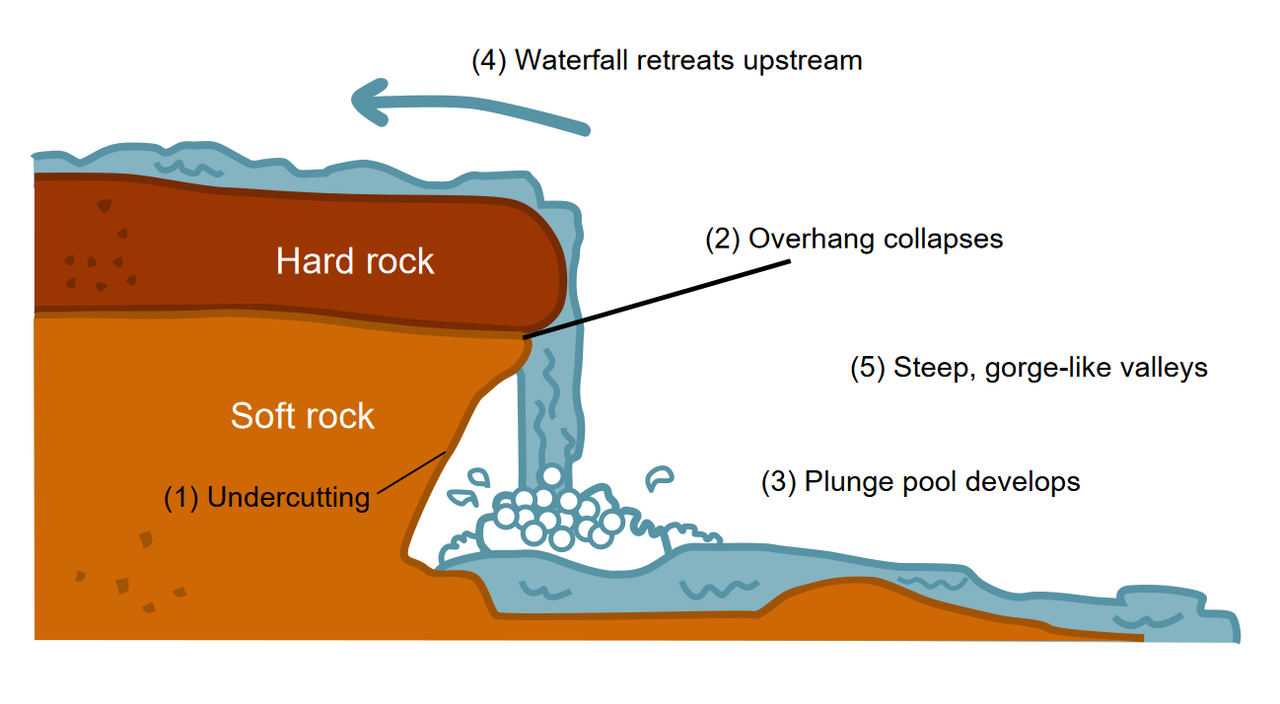}
\end{center}

\medskip
\textbf{Problem: } Draw a geological cross-sectional diagram illustrating the scientific principle of "Waterfall formation and retreat".\\\\\textbf{Required disciplinary entities}:\\Hard rock, Soft rock, Undercutting, Overhang collapses, Plunge pool develops, Waterfall retreats upstream, Steep, gorge-like valleys.\\\\\textbf{Drawing instructions (requiring knowledge invocation)}:\\1. Based on the principle of "Differential erosion" in geoscience, correctly arrange the vertical relationship between the hard rock layer and the soft rock layer in the cross-sectional diagram, and demonstrate the differences in physical morphology of different rock layers after being impacted by water flow.\\2. Accurately label the key dynamic characteristics during the development of the waterfall, including the erosion cavity at the bottom, the top structure that has lost support, and the landform formed at the place where the water flow falls.\\3. According to the evolution laws of fluvial geomorphology, label the overall movement direction of the waterfall over time, as well as the typical valley landform formed as a result.

\medskip
\hrule
\medskip

\noindent\textbf{Criterion 1:} \textit{The diagram must include the required geological units Hard rock and Soft rock, correctly arranged vertically with hard rock over soft rock per the differential erosion principle, showing different physical erosion forms of the two rock layers.}

\begin{itemize}[leftmargin=*,nosep]
    \item \textbf{Point 1.1:} Is Hard rock explicitly depicted as a distinct geologic layer in the diagram?
    \item \textbf{Point 1.2:} Is Soft rock explicitly depicted as a distinct geologic layer in the diagram?
    \item \textbf{Point 1.3:} Is Hard rock correctly positioned vertically above Soft rock in the cross-section profile?
    \item \textbf{Point 1.4:} Does the diagram show clear differences in erosion and physical form between the two rock layers consistent with differential erosion?
    \item \textbf{Point 1.5:} Does the diagram avoid placing Soft rock above Hard rock in the vertical arrangement?
    \item \textbf{Point 1.6:} Does the diagram avoid merging Hard rock and Soft rock into a single undifferentiated geologic unit?
\end{itemize}

\medskip
\noindent\textbf{Criterion 2:} \textit{The diagram must correctly depict and label the Undercutting erosion process, including the erosion cavity formed at the base of the overhanging rock.}

\begin{itemize}[leftmargin=*,nosep]
    \item \textbf{Point 2.1:} Is the Undercutting process explicitly depicted in the correct location (eroding soft rock below the overhanging hard rock layer)?
    \item \textbf{Point 2.2:} Is the eroded cavity from undercutting clearly shown in the diagram?
    \item \textbf{Point 2.3:} Is the text label ``Undercutting'' correctly associated with this process?
    \item \textbf{Point 2.4:} Does the diagram avoid placing the undercutting cavity above the hard rock layer?
    \item \textbf{Point 2.5:} Does the diagram avoid failing to show the eroded hollow shape characteristic of undercutting?
\end{itemize}

\medskip
\noindent\textbf{Criterion 3:} \textit{The diagram must correctly depict and label the Overhang collapses process of the unsupported overlying hard rock after undercutting.}

\begin{itemize}[leftmargin=*,nosep]
    \item \textbf{Point 3.1:} Is the Overhang collapses process explicitly associated with the overhanging hard rock at the waterfall edge?
    \item \textbf{Point 3.2:} Is the label ``Overhang collapses'' correctly assigned to this process step?
    \item \textbf{Point 3.3:} Is the collapse shown occurring to the unsupported top hard rock after undercutting of the lower soft rock?
    \item \textbf{Point 3.4:} Does the diagram avoid placing overhang collapse upstream of the current undercut waterfall edge?
    \item \textbf{Point 3.5:} Does the diagram avoid misattributing collapse to only the soft rock layer?
\end{itemize}

\medskip
\noindent\textbf{Criterion 4:} \textit{The diagram must correctly depict and label the Plunge pool develops landform formed at the base of the falling waterfall.}

\begin{itemize}[leftmargin=*,nosep]
    \item \textbf{Point 4.1:} Is the developing plunge pool explicitly depicted at the base where the waterfall drops to the lower river level?
    \item \textbf{Point 4.2:} Is the label ``Plunge pool develops'' correctly associated with this landform?
    \item \textbf{Point 4.3:} Is the plunge pool positioned correctly downstream of the undercut waterfall face?
    \item \textbf{Point 4.4:} Does the diagram avoid placing the plunge pool above the waterfall face in the upper river section?
    \item \textbf{Point 4.5:} Does the diagram avoid omitting the characteristic depressed pool shape at the base of the waterfall?
\end{itemize}

\medskip
\noindent\textbf{Criterion 5:} \textit{The diagram must correctly depict and label the Waterfall retreats upstream direction of waterfall movement over time.}

\begin{itemize}[leftmargin=*,nosep]
    \item \textbf{Point 5.1:} Is the overall upstream retreat direction of the waterfall explicitly shown with an arrow indicator?
    \item \textbf{Point 5.2:} Is the label ``Waterfall retreats upstream'' correctly associated with the direction indicator?
    \item \textbf{Point 5.3:} Does the arrow point in the correct upstream direction toward the original elevated rock headwall?
    \item \textbf{Point 5.4:} Does the diagram avoid showing retreat direction pointing downstream away from the original headwall?
    \item \textbf{Point 5.5:} Does the diagram avoid failing to indicate the direction of retreat over time?
\end{itemize}

\medskip
\noindent\textbf{Criterion 6:} \textit{The diagram must correctly depict and label the Steep, gorge-like valleys formed as a result of repeated waterfall retreat upstream.}

\begin{itemize}[leftmargin=*,nosep]
    \item \textbf{Point 6.1:} Is the steep, gorge-like valley formed by waterfall retreat explicitly depicted in the diagram?
    \item \textbf{Point 6.2:} Is the label ``Steep, gorge-like valleys'' correctly associated with this resulting landform?
    \item \textbf{Point 6.3:} Is the valley shown with the characteristic steep gorge morphology produced by waterfall retreat?
    \item \textbf{Point 6.4:} Does the diagram avoid depicting the retreat-formed valley as a wide, flat non-gorge shape?
    \item \textbf{Point 6.5:} Does the diagram avoid placing the gorge valley upstream of the current waterfall position?
\end{itemize}

\end{tcolorbox}
\clearpage

\begin{tcolorbox}[title = {Illustrative Examples of rubric generator in Multidisciplinary Reasoning Domain}, breakable, fontupper=\small, fonttitle=\small]

\medskip
\begin{center}
\includegraphics[height=3.2cm]{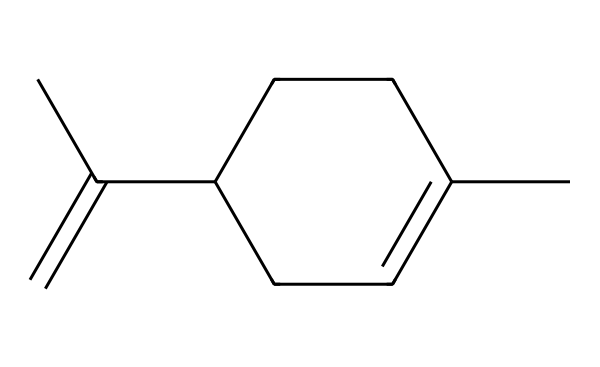}
\end{center}

\medskip
\textbf{Problem: } You are a crew of die-hard concert fans camping overnight, using a citrus-scented repellent to keep bugs at bay. To pass the time, your chemist friend challenges you to draw the active compound's structure from these clues:\\\\1. It is a cyclic molecule with a six-carbon ring containing exactly one carbon-carbon double bond (no other rings).\\2. A methyl group (--CH$_3$) is bonded to one of the two ring carbons involved in that double bond.\\3. Attached to the ring carbon that is three positions away from the methyl-bearing carbon (counting around the ring, moving away from the double bond's other side) is an isopropenyl group---this group has a central carbon connected to both a CH$_3$ and a CH$_2$= (vinyl) group, with that central carbon directly linked to the ring.\\4. Draw the most stable (lowest-energy) conformation of this molecule, considering that the ring's double bond limits some rotations; arrange the groups to minimize steric crowding as much as possible.\\\\What structure do you sketch?

\medskip
\hrule
\medskip

\noindent\textbf{Criterion 1:} \textit{The molecule has exactly one six-carbon ring (no other rings) containing exactly one carbon-carbon double bond within the ring.}

\begin{itemize}[leftmargin=*,nosep]
    \item \textbf{Point 1.1:} Is the molecule depicted with exactly one six-membered carbon ring?
    \item \textbf{Point 1.2:} Does the six-carbon ring contain exactly one carbon-carbon double bond?
    \item \textbf{Point 1.3:} Does the molecule have no additional rings of any size?
    \item \textbf{Point 1.4:} Does the illustration avoid having more than one carbon-carbon double bond in the six-carbon ring?
\end{itemize}

\medskip
\noindent\textbf{Criterion 2:} \textit{A methyl group is bonded to one of the two ring carbons that form the ring's carbon-carbon double bond.}

\begin{itemize}[leftmargin=*,nosep]
    \item \textbf{Point 2.1:} Is a methyl group covalently bonded directly to the six-membered ring?
    \item \textbf{Point 2.2:} Is the methyl group attached to one of the two carbons that form the ring's carbon-carbon double bond?
    \item \textbf{Point 2.3:} Does the illustration avoid attaching the methyl group to a ring carbon that is not part of the ring's double bond?
\end{itemize}

\medskip
\noindent\textbf{Criterion 3:} \textit{A correctly structured isopropenyl group is bonded to the ring carbon that is three positions away from the methyl-bearing carbon (counting around the ring away from the other double bond carbon).}

\begin{itemize}[leftmargin=*,nosep]
    \item \textbf{Point 3.1:} Is the central carbon of the isopropenyl group directly bonded to the target six-membered ring carbon?
    \item \textbf{Point 3.2:} Does the isopropenyl group have the correct structure: central carbon connected to one methyl group and one terminal CH$_2$= double bond?
    \item \textbf{Point 3.3:} Is the isopropenyl group attached to the ring carbon that is three positions away from the methyl-bearing carbon when counting as described?
    \item \textbf{Point 3.4:} Does the illustration avoid attaching the isopropenyl group to any incorrect ring position?
\end{itemize}

\medskip
\noindent\textbf{Criterion 4:} \textit{The structure depicts the most stable (lowest-energy) conformation, with groups arranged to minimize steric crowding accounting for rotation limitations from the ring double bond.}

\begin{itemize}[leftmargin=*,nosep]
    \item \textbf{Point 4.1:} Are the substituents arranged in the conformation that minimizes steric crowding between the groups?
    \item \textbf{Point 4.2:} Does the structure avoid adopting a higher-energy conformation that increases steric hindrance between substituents?
    \item \textbf{Point 4.3:} Does the depicted conformation account for the rotation limitations imposed by the ring's carbon-carbon double bond?
\end{itemize}

\end{tcolorbox}

\end{document}